\documentclass[conference]{IEEEtran}
\usepackage{color}
\usepackage{mathpazo}
\usepackage{times}
\usepackage{color, colortbl}
\definecolor{win}{rgb}{1,1,0}
\usepackage{stfloats}
\usepackage{algorithm}
\usepackage{xcolor}
\usepackage[noend]{algpseudocode}
\usepackage{adjustbox}
\usepackage{enumerate}
\usepackage{array}
\newcommand{\PreserveBackslash}[1]{\let\temp=\\#1\let\\=\temp}
\newcolumntype{C}[1]{>{\PreserveBackslash\centering}p{#1}}
\usepackage{etoolbox}
\makeatletter
\patchcmd{\@makecaption}
  {\scshape}
  {}
  {}
  {}
\makeatother

\usepackage{amsmath, amsthm}
\theoremstyle{plain}
\newtheorem{thm}{Theorem}
\newtheorem{lem}[thm]{Lemma}
\newtheorem{cor}[thm]{Corollary}
\newtheorem{obs}[thm]{Observation}

\newtheorem{assum}[thm]{Assumption}
\theoremstyle{definition}
\newtheorem{defn}{Definition}
\newtheorem{exmp}{Example}

\theoremstyle{remark}

\usepackage{amsfonts}
\usepackage{latexsym}
\usepackage{amssymb,amsbsy}
\usepackage{cite,url}
\usepackage{mathtools}
\usepackage{amscd}

\usepackage{xcolor,stackengine}

\usepackage{upref}
\usepackage{graphicx}
\usepackage{psfrag}
\usepackage{multirow}
\usepackage{ bbold }
\usepackage{mathtools}
\usepackage{xparse}
\DeclarePairedDelimiterX{\set}[1]{\{}{\}}{\setargs{#1}}
\NewDocumentCommand{\setargs}{>{\SplitArgument{1}{;}}m}
{\setargsaux#1}
\NewDocumentCommand{\setargsaux}{mm}
{\IfNoValueTF{#2}{#1} {#1\,\delimsize|\,\mathopen{}#2}}

\DeclareMathOperator{\curl}{\text{Curl}}
\DeclarePairedDelimiter\abs{\lvert}{\rvert}

\DeclareMathOperator*{\argmin}{arg\,min}

\usepackage{tikz}
\usetikzlibrary{positioning}
\pgfdeclarelayer{bg}    
\pgfsetlayers{bg,main}
\definecolor {processblue}{cmyk}{0.96,0,0,0}

\renewcommand{\leq}{\leqslant}

\renewcommand{\geq}{\geqslant}

\newcommand{\given}{\:\vert\:}

\newcommand{\R}{\mathbb{R}}
\newcommand{\E}{\mathbb{E}}
\newcommand{\N}{\mathbb{N}}

\newcommand{\B}{\mathcal{B}}

\newcommand{\Cc}{\overline{\mathcal{C}}}
\newcommand{\C}{\mathcal{C}}

\newcommand{\comment}[1]{}

\renewcommand{\baselinestretch}{0.95}

\newcommand{\Vt}{\tilde{V}}
\newcommand{\vt}{\tilde{v}}
\newcommand{\bt}{\tilde{b}}
\newcommand{\at}{\tilde{a}}
\newcommand{\At}{\tilde{A}}
\newcommand{\Wt}{\tilde{W}}
\newcommand{\wt}{\tilde{w}}
\newcommand{\mt}{\tilde{m}}

\newcommand{\norm}[1]{\left\lVert#1\right\rVert}
\newcommand{\twonorm}[1]{\norm{#1}_{2}}

\newcommand{\co}{\text{Co}}
\newcommand{\cco}{\overline{\text{Co}}}
\newcommand{\bo}{\text{Bo}}
\renewcommand\vec{\mathbf}
\newcommand{\uu}[1]{u^{(#1)}}
\newcommand{\YY}[1]{y^{(#1)}}
\newcommand{\Class}{Y}
\newcommand{\Classes}{\mathcal{Y}}

\newcommand{\cU}{\mathcal{U}}
\renewcommand{\l}{\ell}

\tikzset{
    png export/.style={
        external/system call/.add={}{; convert -density 300 -transparent white "\image.pdf" "\image.png"},
        /pgf/images/external info,
        /pgf/images/include external/.code={
            \includegraphics[width=\pgfexternalwidth,height=\pgfexternalheight]{##1.png}
        },
    }
}
\begin{document}
\usetikzlibrary{arrows.meta,automata,positioning, calc}
\title{Expert Graphs: Synthesizing New Expertise via Collaboration}

\IEEEoverridecommandlockouts

\author{
 \IEEEauthorblockN{Bijan Mazaheri\IEEEauthorrefmark{2}, Siddharth Jain\IEEEauthorrefmark{1}, and Jehoshua Bruck\IEEEauthorrefmark{1}}\\
\IEEEauthorblockA{\IEEEauthorrefmark{1}Electrical Engineering,
California Institute of Technology, U.S.A.
\texttt{\{sidjain,bruck\}@caltech.edu}}\\
\IEEEauthorblockA{\IEEEauthorrefmark{2}Computing \& Mathematical Sciences, California Institute of Technology, U.S.A.
\texttt{bmazaher@caltech.edu}}
\thanks{}
}

\maketitle

\begin{abstract}
Consider multiple experts with overlapping expertise working on a classification problem under uncertain input. What constitutes a consistent set of opinions? How can we predict the opinions of experts on missing sub-domains? In this paper, we define a framework of to analyze this problem, termed ``expert graphs.'' In an expert graph, vertices represent classes and edges represent binary opinions on the topics of their vertices. We derive necessary conditions for expert graph validity and use them to create ``synthetic experts'' which describe opinions  consistent with the observed opinions of other experts. We show this framework to be equivalent to the well-studied linear ordering polytope. We show our conditions are not sufficient for describing all expert graphs on cliques, but are sufficient for cycles.
\end{abstract}
\section{Introduction}\label{sec:intro}
The merits of specialization have distributed human knowledge across many overlapping silos. Accessing this network of human knowledge requires soliciting opinions from experts with different overlapping expertise - a daunting experience fraught with inconsistency and confusion. This paper will attempt to explain how this perceived inconsistency can emerge from uncertain inputs. From this understanding, we will establish new rules for ``consistency,'' which we can use to synthesize opinions in areas of expertise that we may not have access to.

We will illustrate our setting with an example from medicine. Consider a patient suffering from a cough who is deciding what kind of doctor to see. There may be a number of reasons for the cough including pollen allergies, a virus, or perhaps cancer. In addition, we will consider uncertainty in the form of random variable input. For example, our patient may not know whether he or she has been exposed to asbestos -- a significant risk factor in lung cancer. Each state of this random variable can be associated with a \emph{categorical distribution} given by a vector of probabilities for each disease. For example,
\begin{center}
\setlength{\tabcolsep}{0pt}
\scalebox{1}{\begin{tabular}{|C{2cm}|C{2cm}|C{2cm}|C{2cm}|}
        \hline
         &  Pollen & Virus  & Cancer\\
        \hline
        Asbestos & .2 & .1 & .7\\
        \hline
        No Asbestos & .2 & .5 & .3 \\
        \hline
    \end{tabular}}
\end{center}
Different specialists exist for subsets of these categories: a general practitioner may be an expert at distinguishing allergies from viruses, an oncologist may have experience with viruses and cancer, and an expert on environmental exposure may know the difference between pollen-induced allergies and toxin-induced cancers. 

Since none of these experts have studied all three causes, the probabilities in the above table are invisible to them. Instead, our experts have formed \emph{situational opinions}, consisting of conditional probabilities. For example, the oncologist may have a sense of the percentage of her asbestos-exposed patients who are ultimately diagnosed with cancer, which is $\frac{7}{8}$. Similarly, she understands that only $\frac{3}{8}$ of her unexposed patients develop cancer. Together, she may report an average of these situational opinions based on the perceived probabilities of asbestos exposure. This calculation may be unconscious or inaccessible (as is the case with many machine learning algorithms). Hence, we may only have access to the average of these situational opinions.

To see how inconsistencies can develop, consider Table~\ref{tab:nontransitivity}.
\begin{table}
\caption{A table of probabilities for three possible input states $\uu{1}, \uu{2}, \uu{3}$ and three possible classes $\YY{1}, \YY{2}, \YY{3}$ which yields nontransitive preferences.}
    \begin{center}
    \setlength{\tabcolsep}{0pt}
    \scalebox{1}{\begin{tabular}{|C{1cm}|C{2cm}|C{2cm}|C{2cm}|}
        \hline
         & $\YY{1}$ & $\YY{2}$ & $\YY{3}$\\
        \hline
        $\uu{1}$ & $.01$ & $.90$ & $.09$ \\
        \hline
        $\uu{2}$ & $.09$ & $.01$ & $.90$ \\
        \hline
        $\uu{3}$ & $.90$ & $.09$ & $.01$\\
        \hline
    \end{tabular}}
    \end{center}
    \label{tab:nontransitivity}
\end{table}
Each column consists of one small ($.01$), one medium ($.09$) and one large ($.90$) probability, each of which dominates the previous probability in this order. Notice that the $\YY{1}$ column dominates the $\YY{2}$ column at $\uu{2}$ and $\uu{3}$. Similarly, $\YY{2}$ dominates the $\YY{3}$ at $\uu{1},\uu{3}$, and $\YY{3}$ dominates $\YY{1}$ at $\uu{1},\uu{2}$. Due to the cyclic rotation of probabilities, the average opinions of all three possible experts return the same value:
\begin{equation*}
    \frac{1}{3}\left(\frac{.01}{.91} + \frac{.09}{.10} + \frac{.90}{.99}\right) > .6
\end{equation*}
Hence, this table yields a cycle of preference ($\YY{1} \leftarrow \YY{2} \leftarrow \YY{3} \leftarrow \YY{1}$) despite having perfectly knowledgeable experts. No matter which class is chosen, one expert will vehemently assert that there is at least a $60\%$ chance we are wrong! This is an example of what is known in voting theory as the Condorcet Paradox \cite{Condorcet}, which arises when voter preferences are arranged in a similarly cyclic fashion.

The unintuitive nature of combining expertise in the presence of uncertainty calls for a careful treatment of this problem. This paper will develop a framework for understanding networks of experts with and overlapping knowledge, called \emph{expert graphs}.
 
\subsection{Related work}

In voting theory, there is significant work on the `linear ordering polytope''  which corresponds to the set of possible networks of hypothetical pairwise elections given a population of rankings \cite{fishburn}. This field has separately developed what we call the ``curl condition'' and shown that it is necessary and sufficient for $n \leq 5$ classes, and just necessary for $n \geq 6$ classes. In this setting, each voter can be thought of as a ``state'' of uncertainty with an absolute ranking. Our problem introduces aleatoric uncertainty to this setting, resulting in soft preferences instead of binary rankings.

The task of synthesizing our experts' knowledge is essentially that of a classifier built from already-trained pairwise classifiers. There are multiple frameworks that exist to use the outcome of pairwise comparisons to obtain multiclassifiers (see \cite{ Dietterich,Singer}). The focus of this literature is on the \emph{design} of classifiers for aggregation. We will instead consider the setting of \emph{unengineered} binary classifiers, referred to as experts.

Related literature also exists in ensemble methods \cite{ensemblemethods, boosting, bagging}, which generally focus on combining experts trained on the \emph{same} task, and whose knowledge is \emph{imperfect} due to differing training datasets. In contrast, our experts are perfect, but trained on \emph{different} tasks.

\subsection{Paper outline}
\begin{itemize}
    \item In Section~\ref{sec:expert_graphs} we establish notation and develop the \emph{expert graph} framework.
    \item In Section~\ref{sec:curl_condition} we define the \emph{curl} and \emph{curl consistent weighted digraphs} (CCWDs). We prove this condition is necessary for the set of feasible expert graphs.
    \item In Section~\ref{sec:synthetic_experts} we use the curl condition to give curl consistency bounds on missing opinions. We call these edges \emph{synthetic experts}.
    \item In Section~\ref{sec:equivalence} we discuss a similar framework, linear ordering graphs (LOGs). We prove this framework is equivalent to the expert graph framework. We use this equivalence to prove when the curl condition is sufficient (or insufficient) for describing all possible expert graphs in various settings.
    \item In Section~\ref{sec:conclude} we conclude the paper and point to exciting future directions and applications.
\end{itemize}
\section{The expert graph framework}\label{sec:expert_graphs}
\subsection{Notation}
\newcommand{\onenorm}[1]{\norm{#1}_1}
\newcommand{\indicator}[1]{\mathbb{1}[#1]}
\newcommand{\indvec}[1]{\vec{1_{\l}}[#1]}
\newcommand{\onevec}{\vec{1_\l}}
\newcommand{\simplex}{\triangle}
\newcommand{\sit}{G^{\vec{P}}_u}
\newcommand{\reg}{G^{\vec{P}}_d}

We will first provide notation used throughout the paper. 
\begin{itemize}
    \item $[\ell]$ is used to denote the set $\{1, 2, \ldots,\ell\}$ for any $\ell \in \N$.
    \item$|A|$ denotes the cardinality of set $A$.
    \item $\indicator{c}$ will be used for an indicator function which is $1$ if condition $c$ is met and $0$ otherwise.
    \item Any bold symbol unless otherwise stated will be used to denote a vector.  $\boldsymbol{v} = (v_1, \ldots,v_{\ell})^T$ denotes a $\ell$-length vector. To keep track of indexing more easily, we will generally use superscripts to index class and subscripts to index state. For example $p_u^{(i)} = \Pr(\YY{i} \given u)$, $\vec{p}_u = (p_u^{(1)}, \ldots,  p_u^{(n)})^\top$ and $\vec{p}^{(i)} = (p_{u^{(1)}}^{(i)}, \ldots,  p_{u^{(k)}}^{(i)})^\top$.
    \item $\onevec$ denotes an all $1$ vector of size $\ell$.
    \item $\triangle^\l$ will be used to denote vectors of size $\l$ which can are probability distributions. That is, $\lambda \in \triangle^\l$ iff $\lambda \in [0, 1]^\l$ and $\vec{1}^\top \lambda = 1$.
    \item We use $\prec$, $\succ$, $\preceq$, $\succeq$ to denote element-wise inequality. For example, we say $\vec{w} \succeq \vec{v}$ if $w_i\geq v_i ~\forall~ i\in[\ell].$
    \item The L1 norm of a vector $\boldsymbol{v}$ is given by 
$\onenorm{\boldsymbol{v}} = \sum_{i=1}^{\ell}|v_i|$ and the L2 norm is given by $\|\boldsymbol{v}\|_2 = \sqrt{\sum_{i=1}^{\ell}|v_i|^2}$.
    \item We will use $\co(S)$ to denote the open convex hull of $S$, $\cco(S)$ to denote the closed convex hull, and $\bo(\cdot)$ to denote the boundary.
    \item We will use the convention of lowercase letters denoting specific assignments to random variables. For example, $\Pr(A \given u) = \Pr(A \given U = u)$.
\end{itemize}
\subsection{Expert graphs}
\newcommand{\EE}{E}

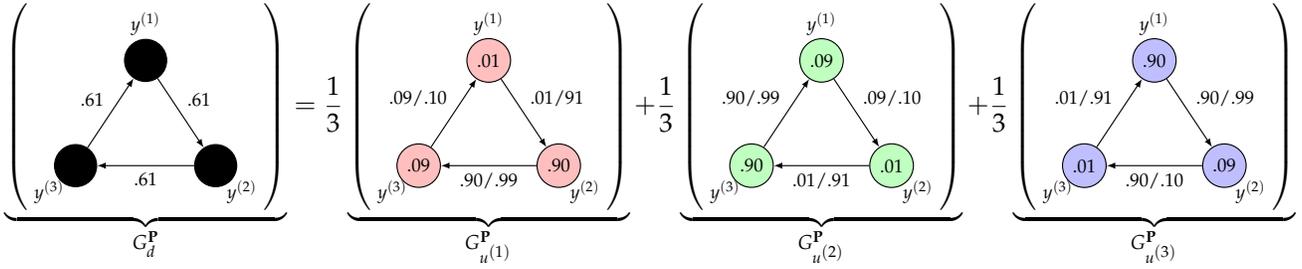
\begin{figure*}
\centering
\begin{center}
    \setlength{\tabcolsep}{0pt}
    \scalebox{1}{\begin{tabular}{|C{1cm}|C{2cm}|C{2cm}|C{2cm}|}
        \hline
         & $\YY{1}$ & $\YY{2}$ & $\YY{3}$\\
        \hline
        \rowcolor{red!25} $\uu{1}$ & $.01$ & $.90$ & $.09$ \\
        \hline
        \rowcolor{green!25} $\uu{2}$ & $.09$ & $.01$ & $.90$ \\
        \hline
        \rowcolor{blue!25}$\uu{3}$ & $.90$ & $.09$ & $.01$\\
        \hline
    \end{tabular}}
\end{center}
\begin{equation*}
    \underbrace{\left(\scalebox{.7}{\begin {tikzpicture}[-latex, auto,node distance =1cm and 1.33 cm ,on grid,
semithick, baseline, anchor=base, state/.style ={ circle, draw, minimum width =.8 cm}]
\node (center) {};
\node[state, fill] (c) [below left = of center]{$.2$};
\node (clab) [below left = .5cm and .5cm of c]{$\YY{3}$};
\node[state, fill] (a) [above = of center]{$.5$};
\node (alab) [above = .7cm of a]{$\YY{1}$};
\node[state, fill] (b) [below right = of center]{$.3$};
\node (blab) [below right = .5cm and .5cm of b]{$\YY{2}$};
\path (a) edge node[above right] {$.61$} (b);
\path (b) edge node[below] {$.61$} (c);
\path (c) edge node[above left] {$.61$} (a);
\end{tikzpicture}}\right)}_{\reg} = \frac{1}{3}\underbrace{\left(\scalebox{.7}{\begin {tikzpicture}[-latex, auto,node distance =1cm and 1.33 cm ,on grid,
semithick, baseline, anchor=base, state/.style ={ circle, draw, minimum width =.8 cm}]
\node (center) {};
\node[state, fill=red!25] (c) [below left = of center]{$.09$};
\node (clab) [below left = .5cm and .5cm of c]{$\YY{3}$};
\node[state, fill=red!25] (a) [above = of center]{$.01$};
\node (alab) [above = .7cm of a]{$\YY{1}$};
\node[state, fill=red!25] (b) [below right = of center]{$.90$};
\node (blab) [below right = .5cm and .5cm of b]{$\YY{2}$};
\path (a) edge node[above right] {$.01/91$} (b);
\path (b) edge node[below] {$.90/.99$} (c);
\path (c) edge node[above left] {$.09/.10$} (a);
\end{tikzpicture}}\right)}_{G_{\uu{1}}^{\vec{P}}} + \frac{1}{3}\underbrace{\left(\scalebox{.7}{\begin {tikzpicture}[-latex, auto,node distance =1cm and 1.33 cm ,on grid,
semithick, baseline, anchor=base, state/.style ={ circle, draw, minimum width =.8 cm}]
\node (center) {};
\node[state, fill=green!25] (c) [below left = of center]{$.90$};
\node (clab) [below left = .5cm and .5cm of c]{$\YY{3}$};
\node[state, fill=green!25] (a) [above = of center]{$.09$};
\node (alab) [above = .7cm of a]{$\YY{1}$};
\node[state, fill=green!25] (b) [below right = of center]{$.01$};
\node (blab) [below right = .5cm and .5cm of b]{$\YY{2}$};
\path (a) edge node[above right] {$.09/.10$} (b);
\path (b) edge node[below] {$.01/.91$} (c);
\path (c) edge node[above left] {$.90/.99$} (a);
\end{tikzpicture}}\right)}_{G_{\uu{2}}^{\vec{P}}} + \frac{1}{3}\underbrace{\left(\scalebox{.7}{\begin {tikzpicture}[-latex, auto,node distance =1cm and 1.33 cm ,on grid,
semithick, baseline, anchor=base, state/.style ={ circle, draw, minimum width =.8 cm}]
\node (center) {};
\node[state, fill=blue!25] (c) [below left = of center]{$.01$};
\node (clab) [below left = .5cm and .5cm of c]{$\YY{3}$};
\node[state, fill=blue!25] (a) [above = of center]{$.90$};
\node (alab) [above = .7cm of a]{$\YY{1}$};
\node[state, fill=blue!25] (b) [below right = of center]{$.09$};
\node (blab) [below right = .5cm and .5cm of b]{$\YY{2}$};
\path (a) edge node[above right] {$.90/.99$} (b);
\path (b) edge node[below] {$.90/.10$} (c);
\path (c) edge node[above left] {$.01/.91$} (a);
\end{tikzpicture}}\right)}_{G_{\uu{3}}^{\vec{P}}}
\end{equation*}
\caption{The expert graph $G^{\vec{P}}_d$ given by situational expert graphs with $\vec{P}$ given by Table~\ref{tab:nontransitivity} (displayed again here) and $d(\cdot)$ the uniform distribution. Probabilities from Table~\ref{tab:nontransitivity} are given inside vertices of situational expert graphs, and expert opinions $f_u(\YY{i}, \YY{j})$ are given as weights on edge $\YY{i} \rightarrow \YY{j}$. The expert graph $\reg$ has edge-weights averaged over the situational expert graphs $G_{\uu{1}}^{\vec{P}}, G_{\uu{2}}^{\vec{P}}, G_{\uu{3}}^{\vec{P}}$.}
\label{fig:situational_decomp}
\end{figure*}

We consider a prediction setting with:
\begin{itemize}
    \item $X$: Known and understood input.
    \item $U$: Unknown, but understood input.
    \item $Y$: Labels of a classification problem.
\end{itemize}
This setting will contain two types of uncertainty. In our introductory example, we were uncertain of exposure to asbestos, which was relevant to our prediction task. We represent uncertain input by a random variable $U$ that takes values in alphabet $\cU = [k]$ with probability distribution $d(\cdot): \cU \mapsto \triangle^{k}$.

Even given full knowledge of asbestos exposure and other factors ($U$ and $X$), we were not completely sure of the disease. Hence, our labels $\Classes$ are also represented by a random variable partially determined by $U$. Specifically, each label $\Class$ will have a categorical distribution over alphabet $\Classes = \{\YY{1}, \YY{2}, \ldots, \YY{n}\}$ given input $u$:
\begin{align}
    p_u^{(i)} = \Pr(Y = \YY{i} \given u, x)
\end{align}
\begin{assum}\label{assum:0}
We assume for any $u \in \mathcal{U}$, $\vec{p}_u \succ 0$.
\end{assum}
We refer to these $n \times k$ probabilities as the ``probability table,'' $\vec{P}$ (see Table~\ref{tab:nontransitivity} for an example) and denote the vector of probabilities for a given input as $\vec{p}_u \in \simplex^n$. For the rest of the paper we will be considering the probabilities conditioned on the known input $x$, but we will omit the $x$ from the conditioning for simplicity.

An oncologist's perception of probabilities will be based on the patients they see and study daily - a population that is unlikely to have many patients only suffering from allergies. Hence, our experts do not have access to the probability table; they only understand the relative probabilities between the pair of classes in their \emph{area of expertise}. For simplicity, we will consider a list of $m$ \emph{binary} experts $\EE \subseteq \Classes\times \Classes$.
\begin{defn} Expert $(\YY{i}, \YY{j}) \in E$ has access to \textbf{situational opinions} $f_u(\YY{i}, \YY{j})$ for each $u \in \cU$ given by the following conditional probability:
\begin{equation}
\begin{aligned}
    f_u(\YY{i}, \YY{j}) &:= \Pr(\Class = \YY{i} \given \Class \in \{\YY{i}, \YY{j}\}, u)\\
    &= \frac{p_u^{(i)}}{p_u^{(i)} + p_u^{(j)}}
\end{aligned}
\end{equation}
\end{defn}
\begin{obs}
From assumption \ref{assum:0}, $f_u(e)\in (0,1)~\forall~e \in E$.
\end{obs}
\begin{obs}\label{obs:edgeflipping}
    $f_u(\YY{i}, \YY{j}) = 1 - f_u(\YY{j}, \YY{i}).$
\end{obs}
With this we can define useful building blocks, called \emph{situational expert graphs}.
\begin{defn}
A \textbf{situational expert graph} $G^{\vec{p}}_u = (\Classes, E, f_u(\cdot))$ encodes experts' pairwise situational opinions $f_u(e)$ as weights on directed edges $e = (\YY{i},\YY{j}) \in E$.
\end{defn}

In our introductory example, our experts reported average opinions over likely values of $u$. To make this precise, we will study the \emph{overall opinion} of experts over distribution $d(\cdot)$:
\begin{defn}
Expert $e = (\YY{i}, \YY{j})$ will report their \textbf{overall opinion}, the expected value\footnote{In this paper we focus on discrete distributions for which the expectation is a sum. An extension to non-discrete distributions is naturally given by using an integral for the expectation.} of situational opinions:
\begin{equation*}
    \E_d[f_u(e)] = \sum_{u \in \cU} d(u) f_u(e).
\end{equation*}
\end{defn}
Expert graphs will be the analog of situational expert graphs with weights given by \emph{overall} opinions:
\begin{defn}
An \textbf{expert graph} $G^{\vec{P}}_d = (\Classes, E, f_{d}(\cdot))$ encodes experts' pairwise overall opinions $f_d(e) = \E_d(f_u(e))$ as weights on directed edges $e = (\YY{i},\YY{j}) \in E$. An expert graph where $(\Classes, E)$ is a cycle will be referred to as an \textbf{expert cycle}.
\end{defn}
We emphasize that input distribution $d(\cdot)$ and probability table $\vec{P}$ \emph{generate} expert graph $G_d^{\vec{P}}$, but they may not be known to the observer of the graph due to the sometimes inaccessible reasoning of both human and machine classifiers.

\begin{figure}
\centering
\begin{equation*}
\left(\scalebox{.7}{\begin {tikzpicture}[-latex, auto,node distance =.7cm and 1.05cm ,on grid ,
semithick, anchor = base, baseline]
\node[circle,fill,inner sep=0.05pt,label=left:$\YY{4}$] (d){$C_3$};
\node[circle,fill,inner sep=0.05pt,label=above:$\YY{1}$] (a) [above right = of d]{$C_1$};
\node[circle,fill,inner sep=0.05pt,label=right:$\YY{2}$] (b) [below right = of a]{$C_1$};
\node[circle,fill,inner sep=0.05pt,label=below:$\YY{3}$] (c) [below left = of b]{$C_2$};
\path (a) edge [bend right = 0] node[above right] {$0.3$} (b);
\path (b) edge [bend right = 0] node[below right] {$0.3$} (c);
\path (c) edge [bend right = 0] node[below left] {$0.7$} (d);
\path (d) edge [bend right = 0] node[above left] {$0.7$} (a);
\path (a) edge [bend right = 0] node[right] {$0.5$} (c);
\end{tikzpicture}}\right)
=\left(\scalebox{.7}{\begin {tikzpicture}[-latex, auto,node distance =.7 cm and 1.05cm ,on grid ,
semithick, anchor = base, baseline]
\node[circle,fill,inner sep=0.05pt,label=left:$\YY{4}$] (d){$C_3$};
\node[circle,fill,inner sep=0.05pt,label=above:$\YY{1}$] (a) [above right = of d]{$C_1$};
\node[circle,fill,inner sep=0.05pt,label=right:$\YY{2}$] (b) [below right = of a]{$C_1$};
\node[circle,fill,inner sep=0.05pt,label=below:$\YY{3}$] (c) [below left = of b]{$C_2$};
\path (b) edge [bend right = 0] node[above right] {$0.7$} (a);
\path (c) edge [bend right = 0] node[below right] {$0.7$} (b);
\path (c) edge [bend right = 0] node[below left] {$0.7$} (d);
\path (d) edge [bend right = 0] node[above left] {$0.7$} (a);
\path (a) edge [bend right = 0] node[right] {$0.5$} (c);
\end{tikzpicture}}\right)
\end{equation*}
\caption{Two equivalent expert graphs with $\Classes = \{\YY{1}, \YY{2}, \YY{3}, \YY{4}\}$ and $f_d(\YY{1}, \YY{2}) = 0.3$, $f_d(\YY{2},\YY{3}) = 0.3$,
$f_d(\YY{3},\YY{4}) = 0.7$, $f_d(\YY{4},\YY{1}) = 0.7$, $f_d(\YY{1},\YY{3}) = 0.5$.
}
\label{fig:ex_knowledge_graph}
\end{figure}
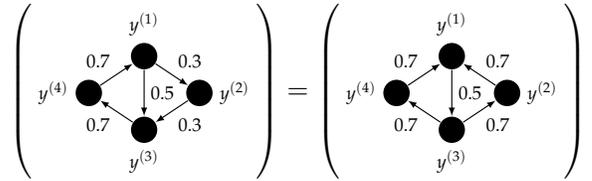

\begin{obs}\label{obs:avgedgeflipping}
    As in Observation~\ref{obs:edgeflipping}, $f_d(\YY{i}, \YY{j}) = 1 - f_d(\YY{j}, \YY{i})$.
\end{obs}
Figure~\ref{fig:ex_knowledge_graph} shows an example of an expert graph with $n=4$ classes and $m=5$ experts and demonstrates the equivalence given in Observation~\ref{obs:avgedgeflipping}.

It will be useful to think of our expert graphs as decompositions of situational expert graphs due to the following observation:
\begin{obs}\label{obs:convexhullexperts}
Let $\mathcal{G}_\cU$ denote the set of all possible situational expert graphs and let $\mathcal{G}$ denote the set of all possible expert graphs. We have that $\mathcal{G} = \cco(\mathcal{G}_\cU)$.
\end{obs}
In this perspective, each situational expert graph can be thought of as representing a specific row of the probability table. An example of decomposing an expert into situational expert graphs is given in Figure~\ref{fig:situational_decomp}.

\section{The curl condition}\label{sec:curl_condition}
\subsection{Definition of curl}
We now give a property on expert graphs called the \emph{curl}, which is motivated by vector calculus and physics.
\begin{defn}
Given an expert graph $G^{\vec{P}}_d = (\Classes, E, f_d)$ and a cycle of indices $\mathcal{C} = (c_1, c_2, \ldots, c_{\ell})$, we define the \textbf{curl} to be:
\begin{equation}\label{eqn:curl_overall}
    \curl(G^{\vec{P}}_d, \mathcal{C}) = f_d(\YY{c_\ell}, \YY{c_1}) + \sum_{i = 1}^{\ell-1} f_d(\YY{c_i}, \YY{c_{i+1}}).
\end{equation}
\end{defn}
\begin{exmp}
In Figure~\ref{fig:ex_knowledge_graph}, $\curl(G, (1, 2, 3)) = .3 + .3 + .5 = 1.1$.
\end{exmp}

It will be useful to also define a similar notion for situational expert graphs, called the \emph{situational curl}.
\begin{defn}
Given a situational expert graph $G^{\vec{P}}_u = (\Classes, E, f_u)$ and a cycle of indices $\C = (c_1, c_2, \ldots, c_{\ell})$, we define the \textbf{situational curl} to be:
\begin{equation}\label{eqn:curl_situational}
    \curl(G^{\vec{P}}_u, \C) =  f_u(\YY{c_\ell}, \YY{c_1}) + \sum_{i = 1}^{\ell-1} f_u(\YY{c_i}, \YY{c_{i+1}}).
\end{equation}
\end{defn}
\begin{obs}\label{obs:reversecycle}
If $\Cc = (c_{\ell},c_{\ell-1},\ldots,c_2,c_1)$ is the reversed direction of the cycle $\mathcal{C}$, applying Observation~\ref{obs:edgeflipping} and Observation~\ref{obs:avgedgeflipping} to every edge gives:
\begin{align*}
    \curl(\sit, \Cc) &= \ell-\curl(G_u,\C),\\
    \curl(\reg, \Cc) &= \ell-\curl(G_d,\C).
\end{align*}
\end{obs}

\subsection{The curl condition}\label{subsec:curlcond}

We will now give upper and lower bounds on the curl. We will first give this condition for \emph{situational expert graphs} which will then extend to regular expert graphs by linearity of expectation.
\begin{lem}
\label{lem:boundcurl}
Given a situational expert graph $G^{\vec{P}}_u$ cycle $\C$ of length $\ell \geq 3$, the situational curl must follow:
\begin{equation}
    1 < \curl(\sit, \C) < \ell-1.
\end{equation}
\end{lem}

\begin{IEEEproof}
We begin by showing the lower bound. Recall that 
    \begin{equation}
        \curl(\sit, \C) = \frac{p_u^{c_\ell}}{p_u^{c_\ell} + p_u^{c_1}} + \sum_{i=1}^{\ell-1} \frac{p_u^{(c_i)}}{p_u^{(c_i)} + p_u^{(c_{i+1})}}.
    \end{equation}
    The denominators of every term are strictly upper bounded by $1$. This gives
    \begin{equation}
        \curl(\sit, \C) > \sum_{i=1}^{\ell} p_u^{(c_i)} = 1.
    \end{equation}
    
Now, we can use this result to prove the upper bound. By Observation~\ref{obs:reversecycle} we have
\begin{align*}
    \curl(\sit, \Cc) = \ell - \curl(\sit, \C) \stackrel{(a)}< \ell - 1,
\end{align*}
where (a) is given by invoking the lower bound proved above.
\end{IEEEproof}

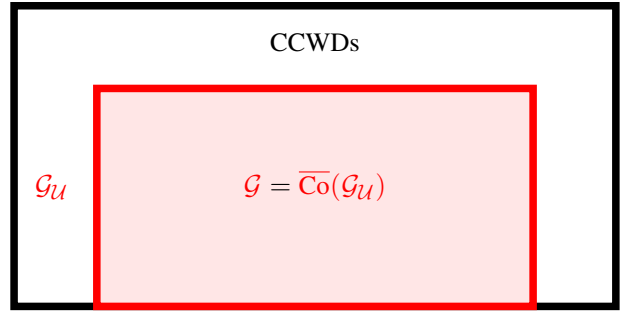
\begin{figure}
\centering
\begin {tikzpicture}[-latex, auto,node distance =2cm and 2cm ,on grid ,
semithick, baseline, anchor=base]
\draw [line width=.1cm] (-4,-2) rectangle (4, 2);
\filldraw[line width=.1cm, color=red, fill = red!10] (-2.9,-2) rectangle (2.9, .9);
\node at (0, -0.5){ $\color{red} \mathcal{G} \color{black} = \color{red}\cco(\mathcal{G}_\cU)$};
\node[color = red] at (-3.5, -.5) { $\mathcal{G}_\cU$};
\node at (0, 1.4) {CCWDs};
\end{tikzpicture}
\caption{$\color{red}\mathcal{G}_\cU\color{black}$ represents the set of possible situational expert graphs given by some probability vector, and $\color{red}\mathcal{G}\color{black}$ is their closed convex hull, representing the possible overall expert graphs. The set of curl consistent weighted digraphs (CCWDs) is given in black, fully containing $\color{red}\mathcal{G}\color{black}$ as shown by Lemma~\ref{lem:boundcurl} and Corollary~\ref{cor:curl_bound_d_main}.}
\label{fig:sets_curl}
\end{figure}

\begin{cor}\label{cor:curl_bound_d_main}
Given expert graph $\reg$ and a cycle $\C$ on $\ell \geq 3$ classes, then the curl follows\footnote{The notion of curl can be extended to one vs multiple class experts as well, in which case we can extend the upper bound of corollary \ref{cor:curl_bound_d_main} to $\ell-q+1$, where $q$ represents the total number of classes used by the expert.}
\begin{equation}
    1 < \curl(\reg, \C) < \ell-1.
\end{equation}
\end{cor}
\begin{IEEEproof}
This result follows from Lemma \ref{lem:boundcurl} and linearity of expectation on the definition of $f_d(\cdot)$. 
\end{IEEEproof} 
\begin{defn}
We call a weighted digraph $G = (\Classes, E, f(\cdot))$ obeying $f(\YY{i}, \YY{j}) = 1 - f(\YY{j}, \YY{i})$ \textbf{curl consistent} if all of its cycles obey the curl condition given by Lemma~\ref{lem:boundcurl} and Corollary~\ref{cor:curl_bound_d_main}. The set of such digraphs is refered to as \textbf{curl consistent weighted digraphs (CCWDs)}.
\end{defn}
It is helpful to think of the class of CCWDs as different from the set of expert graphs, which must be generated by some probability table. Corollary \ref{cor:curl_bound_d_main} gives us a \emph{necessary} condition for expert graphs that is a result of them being generated by a probability table $\vec{P}$ and input distribution $d(\cdot)$. This section has shown CCWDs must contain the set of expert graphs $\mathcal{G}$, as illustrated in  Figure~\ref{fig:sets_curl}.

\section{Synthetic experts}\label{sec:synthetic_experts}
\subsection{Bounding missing edges}
The definition of expert graphs is given on digraphs that are not necessarily complete. If we are given an expert graph with missing edge-weights, many possible probability table and distribution pairs ($\vec{P}$ and $d(\cdot)$) may generate the \emph{given} edge-weights. Each of these possible models gives different values on the \emph{missing} edge-weights. Hence, in the absence of a relevant expert, we would like to give ranges which are consistent with \emph{some} model given by $\vec{P}$ and $d(\cdot)$, i.e. membership in the class of expert graphs $\mathcal{G}$. Unfortunately, this problem is very difficult, as we will see from the reduction of our system to the linear ordering polytope in Section~\ref{sec:equivalence}.

Alternatively, we know from the previous section that a violation of the curl condition signifies that no model can generate the weighted digraph as an expert graph. Hence, a more feasible step towards handling missing experts involves bounding edge-weights so that we maintain the graph's curl consistency.

We begin by giving the notion of path weights.
\begin{defn}\label{def:graph}
Let $\B = (b_1,b_2,\cdots,b_\ell)$ denote ordered indices of classes in expert graph $\reg = (\Classes, E, f_d(\cdot))$ with $(\YY{b_i},\YY{b_{i+1}}) \in E~\forall~i\in[\ell-1]$. The \textbf{weight} of a path is given by:
\begin{equation}
    W(\reg, \B) = \sum_{i=1}^{\ell-1} f_d(\YY{b_i},\YY{b_{i+1}}).
\end{equation}
\end{defn}
The path weight is very similar to the curl, but is defined on a path that does not loop back to its starting point.

\begin{lem}\label{lem:synthetic_experts}
Consider CCWD $G = (\Classes, E, f(\cdot))$, with edge $(\YY{i},\YY{j}) \notin E$.  Let $\B^{(ij)}$ be the set of paths in $G$ beginning at vertex $\YY{i}$ and ending at vertex $\YY{j}$:
\begin{equation*}
    \B^{(ij)} = \{B \;:\; b_1 = i, b_{\abs{B}} = j\}.
\end{equation*}
Then, 
\begin{equation}
    1-\min_{\overline{B} \in \B^{(ji)}}W(G, \overline{B}) < f(\YY{i},\YY{j}) <  \min_{B \in \B^{(ij)}}W(G, B).
\end{equation}
\end{lem}
\begin{IEEEproof}
Consider the $\ell$-length cycle $B = (i, \ldots, j)$ and its reverse $\overline{B} = (j, \ldots, i)$. Corollary \ref{cor:curl_bound_d_main} gives
\begin{equation}\label{eq:curlboundB}
    1 < \curl(G, \overline{B}) < \ell-1.
\end{equation}
Notice that 
\begin{equation*}
    \curl(G, \overline{B}) = W(G, \overline{B}) + f(\YY{i}, \YY{j}).
\end{equation*}
Substituting this into Equation~\ref{eq:curlboundB} gives
\begin{equation*}
    1 -  W(G, \overline{B}) < f(\YY{i}, \YY{j}) < \ell-1 -  W(G, \overline{B}).
\end{equation*}
Now, by Observation~\ref{obs:reversecycle}, we can simplify the upper bound to get
\begin{equation}\label{eq:curlboundsimp}
    1 -  W(G, \overline{B}) < f(\YY{i}, \YY{j}) < W(G, B).
\end{equation}
Recall that $B$ and $\overline{B}$ were chosen without loss of generality. Obtaining the optimal bounds over all paths yields the desired result.
\end{IEEEproof}

\begin{figure}
\centering
\scalebox{.8}{
\begin {tikzpicture}[-latex, auto,node distance =2cm and 2cm ,on grid ,
semithick, baseline, anchor=base]
\node (e) {};
\node[circle,fill,inner sep=0.05pt,label= left:$s$] (S) [left = of e]{$C_0$};
\node[circle,fill,inner sep=0.05pt,label= above:$\YY{1}$] (C1) [above = of e]{$C_1$};
\node[circle,fill,inner sep=0.05pt,label= right:$t$] (T) [right = of e]{$C_1$};
\node[circle,fill,inner sep=0.05pt,label=below :$\YY{2}$] (C2) [below = of e]{$C_3$};
\path (S) edge [ red] node[above left] {$0.3$} (C1);
\path (C1) edge [ red] node[above right] {$0.4$} (T);
\path (S) edge [bend left = 15, red, dashed] node[above] {$< 0.7$} (T);
\path (S) edge [bend right = 15, blue, dashed] node[below] {$>0.1$} (T);
\path (T) edge [blue] node[below right] {$0.5$} (C2);
\path (C2) edge [blue] node[below left] {$0.4$} (S);
\end{tikzpicture}}
\caption{The shortest path from $s \rightarrow t$, given by $\color{red} s \rightarrow \YY{1} \rightarrow t$ gives $f_d(s, t) < 0.7$. The shortest path from $t$ to $s$, given by $\color{blue} t \rightarrow \YY{2} \rightarrow S$ gives $f_d(s, t) > 1 - .9 = .1$. Together, we get $f_d(s, t) \in (0.1, 0.7)$.}
\label{fig:synthetic}
\end{figure}
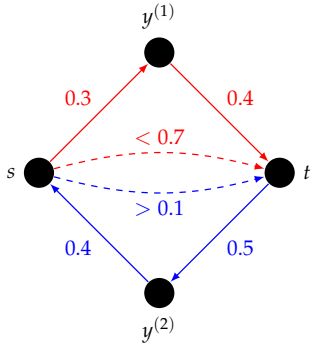

A number of shortest-path algorithms for weighted directed graphs, such as Floyd-Warshall~\cite{Cormen}, can be applied to find the best bounds given by Lemma \ref{lem:synthetic_experts}. An example of this process is illustrated in Figure~\ref{fig:synthetic}.

We will later show that the curl condition is sufficient for describing all possible expert graphs whose edges form a single cycle, giving a case for which these bounds are tight for both CCWDs \emph{and} expert graphs. There are, however, counter-examples of CCWDs that are not achievable by some probability table and input distribution. Thus, in general, these bounds can be improved.

\subsection{$\zeta$-accurate synthetic experts}
We can restrict the possible edge-weights of a synthetic expert to arbitrary precision.
\begin{defn}\label{def:perfectsynthclassifiers}
For $\zeta>0$ call the synthetic expert between $\YY{i}$ and $\YY{j}$ a \textbf{$\zeta$-accurate synthetic expert} if for some lower bound $L$, $f(\YY{i},\YY{j}) \in (L, L+\zeta)$.
\end{defn}
\begin{lem}
\label{lem:perfect_synthetic_classifiers}
Given CCWD $G = (\Classes, E, f(\cdot))$, the synthetic expert between $\YY{i}$ and $\YY{j}$ is $\zeta$-accurate if and only if there exists a cycle of indices $\C = (c_1, \ldots, c_{\ell})$ with $i,j \in \mathcal{C}$ and $\curl(G, \C) \leq 1 + \zeta$.
\end{lem}
\begin{IEEEproof}
    Choose $\YY{j}$ and $\YY{i}$ wlog and denote the shortest path:
    \begin{equation}\label{eq:shortest_paths_notation}
           S^{(ij)} = \argmin_{B \in \B^{(ij)}}W(G, B).
    \end{equation}
Recall from Lemma~\ref{lem:synthetic_experts} that
     \begin{equation*}
        1- W(G, S^{(ji)}) < f(\YY{i}, \YY{j})< W(G, S^{(ij)}).
    \end{equation*}
Let $L = 1- W(G, S^{(ji)})$. The desired gap between bounds is $W(G, S^{(ij)}) - \left(1- W(G, S^{(ji)})\right) \leq \zeta$, which is true if and only if 
\begin{equation*}
    W(G, S^{(ji)}) + W(G, S^{(ij)}) \leq 1 + \zeta.
\end{equation*}
Now consider the cycle $\C^{(iji)} = S^{(ij)} S^{(ji)}$, which is the shortest cycle through $\YY{i}$ and $\YY{j}$. We have
    \begin{equation*}
        \curl(G, \C^{(iji)}) = W(G, S^{(ji)}) + W(G, S^{(ij)}) \leq 1 + \zeta.
    \end{equation*}
\end{IEEEproof}

$\zeta$-accurate synthetic experts describe convergence to fully determining a missing opinion without access to the relevant expert. These synthetic experts can form subgraphs of any CCWD.
\begin{lem}\label{lem:perfect_synthetic_classifier_graphs}
Given a CCWD $G = (\Classes, E, f(\cdot))$, there exists another weighted digraph $G'= (\Classes\cup \Classes', E', f'(\cdot))$ with $\Classes' \cap \Classes= \emptyset$ and $E \cap E' = \emptyset$ such that
\begin{enumerate}
    \item All synthetic experts $e \in E$ in $G'$ are $\zeta$-accurate. That is, we have $f(e) \in (L^{(e)}, L^{(e)} + \zeta)$ for some lower bound $L^{(e)}$.
    \item $G'$ is also curl consistent.
\end{enumerate}
\end{lem}
\begin{IEEEproof}
We give a constructive proof, for which an example is given in Figure~\ref{fig:perfect_synth_expert_graph}. Begin with the set of classes $\Classes$ and $E' = \emptyset$. Now, for each $e = (\YY{i}, \YY{j}) \in E$, do the following:
\begin{enumerate}
    \item Add a vertex $\YY{ij}$ to $\Classes'$ and assign values to $f'(\YY{i}, \YY{ij})$ and $f'(\YY{ij}, \YY{j})$ to create a path from $\YY{i} \rightarrow \YY{ij} \rightarrow \YY{j}$ of weight $f(e) +\frac{\zeta}{2}$.
    \item Similarly add a vertex $\YY{ji}$ to create a path from $\YY{j} \rightarrow \YY{ji} \rightarrow \YY{i}$ of weight $1 - f(e) +\frac{\zeta}{2}$.
\end{enumerate} 
By Lemma~\ref{lem:perfect_synthetic_classifiers}, $e$ is now a $\zeta$-accurate synthetic classifier in $G'$ that includes value $f(e)$.

If this process created a cycle $\C'$ in $G'$ with $\curl(G', \C') \leq 1$, then there must also be a cycle through $G$ with $\curl(G, \mathcal{C}) \leq 1$, which is a contradiction. Thus, we know $G'$ as constructed is curl consistent.
\end{IEEEproof}
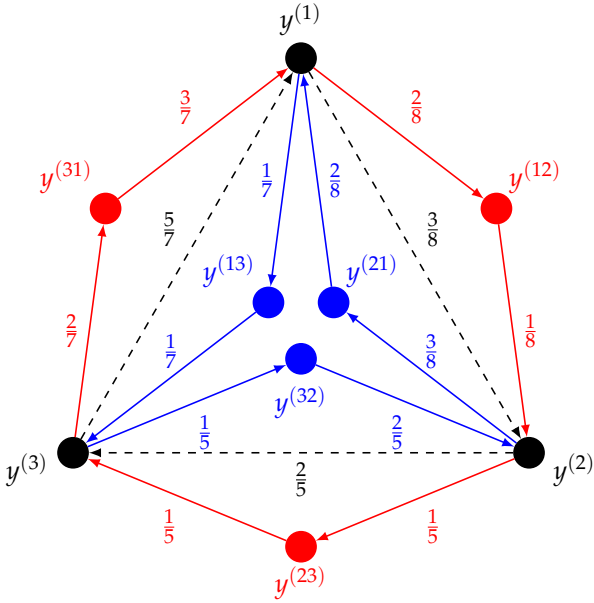
\begin{figure}
    \centering
    \scalebox{1}{\begin {tikzpicture}[-latex, auto,node distance =1cm and 1.33 cm, on grid ,
semithick, baseline, anchor=base, state/.style ={ circle, draw, minimum width =.4 cm, fill}]
\node[state] (A) at (90:3.5) {};
\node[state] (B) at (330:3.5) {};
\node[state] (C) at (210:3.5) {};
\node (Alab) at (90:3.9) {$\YY{1}$};
\node (Blab) at (330:4.2) {$\YY{2}$};
\node (Clab) at (210:4.2) {$\YY{3}$};
\node[state, red] (CAout) at (150:3) {};
\node[red] (CAoutlab) at (150:3.6) {$\YY{31}$};
\node[state, blue] (CAin) at (150:.5) {};
\node[blue] (CAinlab) at (150:1.1) {$\YY{13}$};
\node[state, red] (BCout) at (270:3) {};
\node[red] (BCoutlab) at (270:3.6) {$\YY{23}$};
\node[state, blue] (BCin) at (270:.5) {};
\node[blue] (BCinlab) at (270:1.2) {$\YY{32}$};
\node[state, red] (ABout) at (30:3) {};
\node[red] (ABoutlab) at (30:3.6) {$\YY{12}$};
\node[state, blue] (ABin) at (30:.5) {};
\node[blue] (ABinlab) at (30:1.1) {$\YY{21}$};
\path (A) edge[dashed] node[above right] {$\frac{3}{8}$} (B);
\path (B) edge[dashed] node[below] {$\frac{2}{5}$} (C);
\path (C) edge[dashed] node[above left] {$\frac{5}{7}$} (A);
\path (A) edge[red] node[above right] {$\frac{2}{8}$} (ABout);
\path (ABout) edge[red] node[ right] {$\frac{1}{8}$} (B);
\path (B) edge[blue] node[above] {$\frac{3}{8}$} (ABin);
\path (ABin) edge[blue] node[ right] {$\frac{2}{8}$} (A);
\path (B) edge[red] node[below right] {$\frac{1}{5}$} (BCout);
\path (BCout) edge[red] node[below left] {$\frac{1}{5}$} (C);
\path (C) edge[blue] node[below right] {$\frac{1}{5}$} (BCin);
\path (BCin) edge[blue] node[ below left] {$\frac{2}{5}$} (B);
\path (C) edge[red] node[left] {$\frac{2}{7}$} (CAout);
\path (CAout) edge[red] node[above left] {$\frac{3}{7}$} (A);
\path (A) edge[blue] node[left] {$\frac{1}{7}$} (CAin);
\path (CAin) edge[blue] node[ above] {$\frac{1}{7}$} (C);
\end{tikzpicture}}
    \caption{An example of how to create $\zeta$-accurate synthetic experts that make up any curl consistent graph by adding additional paths. Assume all edge weights given in color have a $+ \frac{\zeta}{4}$. Here, the cycle on $\YY{1}, \YY{2}, \YY{3}$ is created by: (1) Adding shortest paths (shown in \color{red}red \color{black} on the other part of the cycle) with total weight equal to the desired edge weight $f(e) + \frac{\zeta}{2}$. (2) Adding shortest reverse paths (shown in \color{blue}blue \color{black} on the inner part of the cycle) with total weight equal to $1-f(e)+ \frac{\zeta}{2}$.}
    \label{fig:perfect_synth_expert_graph}
\end{figure}

\subsection{Feasibility of networks of synthetic experts}
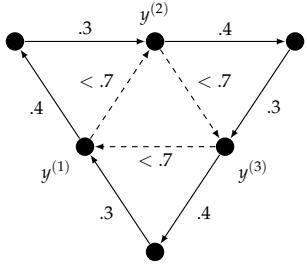
\begin{figure}
    \centering
    \scalebox{.7}{\begin {tikzpicture}[-latex, auto,node distance =1cm and 1.33 cm ,on grid ,
semithick, baseline, anchor=base]
\node (center) {};
\node (cd) [below = 2cm of center] {};
\node (cl) [left = 1.33cm of center] {};
\node (cr) [right = 1.33cm of center] {};
\node[circle,fill,inner sep=0.05pt] (c12) [above left = of cl] {A};
\node[circle,fill,inner sep=0.05pt] (c23) [above right = of cr] {A};
\node[circle,fill,inner sep=0.05pt] (c31) [below = of cd] {A};
\node[circle,fill,inner sep=0.05pt,label=below left:$\YY{1}$] (a) [below left = of center] {A};
\node[circle,fill,inner sep=0.05pt,label=above:$\YY{2}$] (b) [above = of center] {A};
\node[circle,fill,inner sep=0.05pt,label=below right:$\YY{3}$] (c) [below right = of center]{A};
\path (a) edge[dashed] node[above left] {$<.7$} (b);
\path (b) edge[dashed] node[above right] {$<.7$} (c);
\path (c) edge[dashed] node[below] {$<.7$} (a);
\path (c12) edge node[above] {$.3$} (b);
\path (c23) edge node[below right] {$.3$} (c);
\path (c31) edge node[below left] {$.3$} (a);
\path (a) edge node[below left] {$.4$} (c12);
\path (b) edge node[above] {$.4$} (c23);
\path (c) edge node[below right] {$.4$} (c31);
\end{tikzpicture}}
    \caption{An example of how we cannot always achieve any combination of synthetic expert bounds. Here, a choice of just under $.7$ for all the synthetic experts (given in dashed lines) would violate the curl condition on cycle $\YY{1} \rightarrow \YY{2} \rightarrow \YY{3} \rightarrow \YY{1}$.}
    \label{fig:not_all_bound_combos}
\end{figure}

The bounds provided in Lemma~\ref{lem:synthetic_experts} interact with each other when many experts are being synthesized on the same graph. For example, the bounds obtained for each edge may not be attainable simultaneously while still guaranteeing membership in the class of CCWDs (see Figure~\ref{fig:not_all_bound_combos}).

While the bounds for synthetic experts may not be simultaneously attainable, each value within the the bounds does correspond to a valid CCWD.
\begin{lem} \label{lem:at_least_one_feas} Let $S^{(ij)}$ denote the shortest path from $\YY{i}$ to $\YY{j}$ as given in Equation~\ref{eq:shortest_paths_notation}.
Consider a CCWD $G=(\Classes, E, f(\cdot))$ and synthetic experts bounds on $E'$ given by Lemma~\ref{lem:synthetic_experts} and denoted by $f(\YY{i}, \YY{j}) \in (1-S^{(ji)}, S^{(ij)})$. For each $e' = (\YY{i}, \YY{j}) \in E'$ and assignment $v \in (1-S^{(ji)}, S^{(ij)})$, there exists at least one $G^*=(\Classes, E^*, f^*(\cdot))$ with $E^* = E \cup E'$ and $f^*(e') = v$ that is curl consistent.
\end{lem}
\begin{IEEEproof}
We will prove this by inducting on the cardinality of $E'$. First consider the base case of $\abs{E'} = 0$, which is trivially solved by setting $G^* = G$.

Now, assume for induction that the statement holds for all $E'$ with $\abs{E'} \leq q - 1$. Choose $e' = (\YY{i}, \YY{j}) \in E'$  and $v \in (1 - S^{(ji)}, S^{(ij)})$ wlog and use this to define $\Tilde{G} = (\Classes, \Tilde{E}, \Tilde{f}(\cdot))$ where $\Tilde{E} = E \cup \{e'\}$,  $\Tilde{f}(e) = f(e)~\forall~e\in E$ and $\Tilde{f}(e') = v$. 

First, we show that $\Tilde{G}$ is curl consistent. All cycles $\C$ which do not include $e'$ have $\curl(G, \C) = \curl(\Tilde{G}, \C)$. Now consider wlog some cycle $\C^{(ij)} = (c_1, \ldots, i, j, \ldots, c_{\ell-2}) = (i, j) B^{(ji)}$ containing edge $e'$.

Recall from Lemma~\ref{lem:synthetic_experts} that
\begin{equation}\label{eq:recall_synth}
    1 - W(G, S^{(ji)}) = 1-\min_{\overline{B} \in \B^{(ji)}}W(G, \overline{B}) < \Tilde{f}(\YY{i},\YY{j}).
\end{equation}
Now, we can decompose the curl of $\C^{(ij)}$ and apply Equation~\ref{eq:recall_synth} to get
\begin{align*}
    \curl(\tilde{G}, \C^{(ij)}) &= \Tilde{f}(\YY{i},\YY{j}) + W(\tilde{G}, B^{(ji)})\\
    &> 1-\min_{\overline{B} \in \B^{(ji)}}W(G, \overline{B}) + W(\tilde{G}, B^{(ji)}).
\end{align*}
Because $B^{(ji)} \in \B^{(ji)}$, we have
\begin{equation*}
    \curl(\tilde{G}, \C^{(ij)}) > 1.
\end{equation*}
$\C^{(ij)}$ was chosen wlog, so we can conclude that all cycles have curl $>1$. As before, all these cycles can be reversed which means they also have curl upper bounded by $\ell -1$.

So, every assignment to an edge in $E'$ yields a $\tilde{G}$ that is still curl consistent with one more edge. We can now let $E'' = E' \setminus \{e'\}$ and use our inductive assumption on $\Tilde{G}$ with synthetic expert set $E''$.
\end{IEEEproof}

Lemma~\ref{lem:at_least_one_feas} implies that, when referring to the set of CCWDs, the bounds given for synthetic experts are optimal. We will later see that not all curl consistent weighted digraphs can be expert graphs. Recall that Lemma~\ref{lem:perfect_synthetic_classifier_graphs} showed that our bounds could restrict us to being arbitrarily close to \emph{any} CCWD in a network of synthetic experts. As a result, we will see that not all synthetic expert graphs are achievable as an expert graph.

\section{Equivalence to the linear ordering polytope}\label{sec:equivalence}
Proving that the curl condition is sufficient for the set of possible expert graphs $\mathcal{G}$ involves \emph{finding} some probability table $\vec{P}$, and input distribution $d(\cdot)$, which generate $\reg$ for any possible assignment of edge-weights that satisfies the curl condition. In this section we will show that this task is not always possible.

To make the decomposition process easier, we will consider a related set: the ``linear ordering polytope'' \cite{fishburn}. In this setting, situational edges $f_r(\YY{i}, \YY{j})$ are \emph{binary} preferences instead of soft decisions given by conditional probabilities. We will call the analog to a situational expert graph will be called a ``ranking graph.''

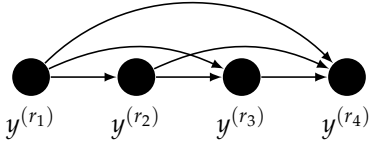
\begin{figure}
    \centering
    \scalebox{1}{\begin {tikzpicture}[-latex, auto,node distance =1.4cm and 1.4cm ,on grid ,
semithick, baseline, anchor=base]
\node[circle, fill, inner sep=0.05pt, label= below:$\YY{r_1}$] (a){$C_0$};
\node[circle,fill,inner sep=0.05pt,label= below:$\YY{r_2}$] (b) [right = of a]{$C_1$};
\node[circle,fill,inner sep=0.05pt,label= below:$\YY{r_3}$] (c) [right = of b]{$C_1$};
\node[circle,fill,inner sep=0.05pt,label= below:$\YY{r_4}$] (d) [right = of c]{$C_3$};
\path (a) edge [bend left = 0]  (b);
\path (b) edge [bend left = 0]  (c);
\path (c) edge [bend left = 0] (d);
\path (a) edge [bend left = 25] (c);
\path (b) edge [bend left = 25] (d);
\path (a) edge [bend left = 45] (d);
\end{tikzpicture}}
    \caption{An example of a ranking graph $H_r(\C, E, f_r)$ with all edge weights $1$. The classes have been arranged by rank from left to right, so that the direction with weight $1$ always points right.}
    \label{fig:ranking_example}
\end{figure}

\begin{defn}
A \textbf{ranking graph} $H_r = (\Classes, E, f_r(\cdot))$ is a complete graph ($E = \Classes\times \Classes$) with a binary edge function $f_r(\cdot): E \mapsto \{0, 1\}$ given by an implicit ordering of indices $r = (r_1, \ldots, r_n)$:
\begin{equation*}
    f_r(\YY{r_i}, \YY{r_j}) = \indicator{i<j}.
\end{equation*}
\end{defn}
An example of a ranking graph on $4$ classes is given in Figure~\ref{fig:ranking_example}.

To distinguish between our frameworks, the analogue of $d(u)$ will be $w(r)$, giving a ``ranking distribution.'' This will yield a component of the linear ordering polytope, which we will call a ``linear ordering graph.''

\begin{defn}
A \textbf{linear ordering graph} (LOG) $H_w = (\Classes, E, f_w(\cdot))$ encodes induced binary probabilities\cite{fishburn} 
\begin{equation}
    f_w(\YY{i}, \YY{j}) = \E_w[f_r(\YY{i}, \YY{j})] = \sum_{r \in \mathcal{R}} w(r) f_r(\YY{i}, \YY{j}),
\end{equation}
where $\mathcal{R}$ represents all $n!$ possible orderings of indices.
\end{defn}
\begin{obs}\label{obs:convexhullrankings}
Let $\mathcal{H_R}$ denote the set of all possible ranking graphs and $\mathcal{H}$ denote the set of all possible LOGs (the linear ordering polytope). Similar to Observation~\ref{obs:convexhullexperts}, we have $\mathcal{H} = \cco(\mathcal{H_R})$.
\end{obs}
\newcommand{\rank}[5]{\begin {tikzpicture}[-latex, auto,node distance =1.4cm and 1.4cm ,on grid ,
semithick, baseline, anchor=base]
\node[circle, draw, fill=#4, minimum width =.4 cm, inner sep=0.05pt, label= below:$\YY{#1}$] (a){};
\node[circle,draw, fill=#4,minimum width =.4 cm, inner sep=0.05pt,label= below:$\YY{#2}$] (b) [right = of a]{};
\node[circle,draw, fill=#4, minimum width =.4 cm, inner sep=0.05pt,label= below:$\YY{#3}$] (c) [right = of b]{};
\node (lab) [above = 1 cm of b] {#5};
\path (a) edge [bend left = 0]  (b);
\path (b) edge [bend left = 0]  (c);
\path (a) edge [bend left = 25] (c);
\end{tikzpicture}}

\newcommand{\cycle}[4]{\scalebox{.8}{\begin {tikzpicture}[-latex, auto,node distance =.75cm and 1 cm ,on grid ,
semithick, baseline, anchor=base, state/.style ={ circle, draw, minimum width =.4 cm, fill}]
\node[state, fill = #4] (A) at (90:1) {};
\node[state, fill = #4] (B) at (330:1) {};
\node[state, fill = #4] (C) at (210:1) {};
\node[] (Alab) at (90:1.4) {$\YY{1}$};
\node[] (Blab) at (330:1.7) {$\YY{2}$};
\node[] (Clab) at (210:1.7) {$\YY{3}$};
\node[] (CAout) at (150:.8) {$#3$};
\node[] (BCout) at (270:.9) {$#2$};
\node[] (ABout) at (30:.8) {$#1$};
\path (A) edge (B);
\path (B) edge (C);
\path (C) edge (A);
\end{tikzpicture}}}

\begin{figure*}
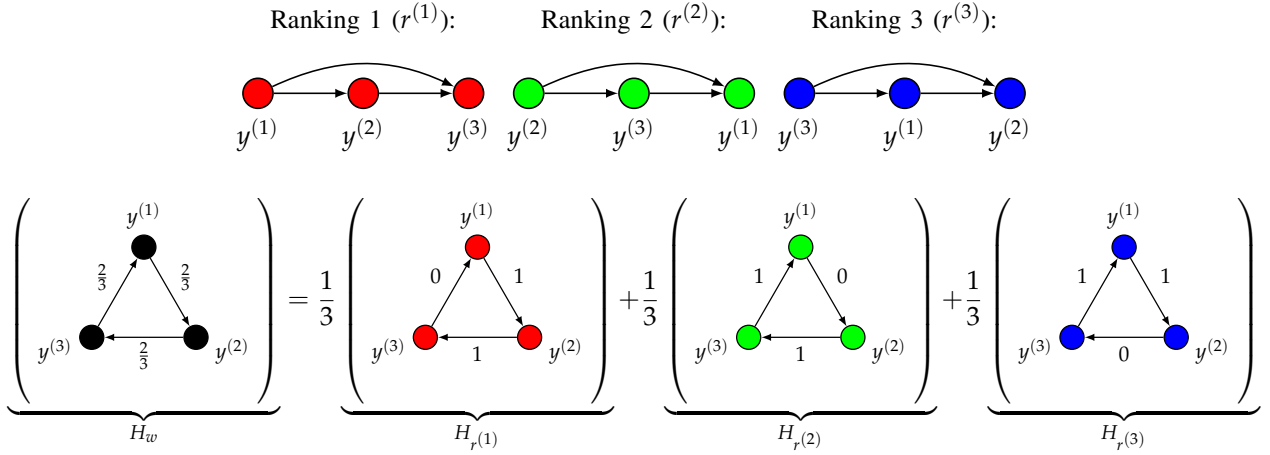

    \centering
\begin{align*}
    \rank{1}{2}{3}{red}{Ranking 1 ($r^{(1)}$):} \rank{2}{3}{1}{green}{Ranking 2 ($r^{(2)}$):}
    \rank{3}{1}{2}{blue}{Ranking 3 ($r^{(3)}$):}
\end{align*}    
    
\begin{equation*}
    \underbrace{\left(\cycle{\frac{2}{3}}{\frac{2}{3}}{\frac{2}{3}}{black}\right)}_{H_w} = \frac{1}{3}\underbrace{\left(\cycle{1}{1}{0}{red}\right)}_{H_{r^{(1)}}} + \frac{1}{3}\underbrace{\left(\cycle{0}{1}{1}{green}\right)}_{H_{r^{(2)}}} + \frac{1}{3}\underbrace{\left(\cycle{1}{0}{1}{blue}\right)}_{H_{r^{(3)}}}
\end{equation*}
    \caption{The LOG $H_w$ given by ranking graphs from rankings $r^{(1)}$, $r^{(2)}$ and $r^{(3)}$ and uniform $w(r) = \frac{1}{3}$. Edge-weights in $H_w$ are averages over the edge-weights in $H_{r^{(1)}}, H_{r^{(2)}}, H_{r^{(3)}}$ determined by the rankings. Decompositions of this form can be used to prove that $H_w \in \co(\mathcal{H_R})= \mathcal{H}$.}
    \label{fig:situational_LOG_decomp}
\end{figure*}
An example of a LOG as a composition of ranking graphs is given in Figure~\ref{fig:situational_LOG_decomp}.

Ranking graphs and LOGs have a condition very close to the curl condition given in Lemma~\ref{lem:boundcurl} and Corollary~\ref{cor:curl_bound_d_main}, with the only difference being a nonstrict inequality.
\begin{lem}\label{lem:RankCurl}
Given a ranking graph $H_r = (\Classes, E, f_r(\cdot))$ and a cycle $\C$ on $\ell \geq 3$ classes, then
    \begin{equation}
        1\leq \curl(H_r,\C) \leq \ell-1.
    \end{equation}
\end{lem}
\begin{IEEEproof}
 Let $\rho(c_i)$ give the ranking of $c_i$ such that $r_{\rho(c_i)} = c_i$. First, note that with $f_r(e) \in \{0, 1\}$ for all $e \in E$, $\curl(H_r, \C)$ must be an integer. Now, assume for contradiction that $\curl(H_r,\C) = 0$, hence $f_r(\YY{c_i}, \YY{c_{i+1}}) = 0~\forall~i \in [\ell-1]$. Recall that $f_r(\YY{r_i}, \YY{r_j}) = \indicator{i<j}$. This implies
 \begin{equation*}
     r_{\rho(c_1)} > r_{\rho(c_2)} > \cdots > r_{\rho(c_{\ell})}.
 \end{equation*}
 A contradiction of transitivity now arises, because we also have $f_r(\YY{c_{\ell}}, \YY{c_1}) = 0$, so $r_{\rho(c_1)} < r_{\rho(c_{\ell})}$.
 
 So we have $\curl(H_r,\C) > 0$. This gives $\curl(H_r,\C) \geq 1$ because we must have integer curl. As with expert graphs, the upper bound follows from considering the curl on the reverse cycle $\Cc$, which must also follow this lower bound.
\end{IEEEproof}
Linearity gives the corollary for the convex hull.
\begin{cor}\label{cor:LOGCurl}
Given LOG $H_w = (\Classes, E, f_w(\cdot))$ and a cycle $\C$ on $\ell \geq 3$ classes, then
\begin{equation}
    1\leq \curl(H_w,\C) \leq \ell-1.
\end{equation}
\end{cor}

Somewhat surprisingly, these two frameworks are essentially\footnote{LOGs may have $\curl(H_w, \C)=1$ or $\curl(H_w, \C)=\ell - 1$ for cycle $\C$ length $\ell$, which violates the strict inequalities in Corollary~\ref{cor:curl_bound_d_main}. To handle this detail, we consider the interior of the linear ordering polytope.} equivalent: any expert graph can also be achieved as a LOG, and any LOG in the interior of the linear ordering polytope can be achieved as an expert graph.

\begin{thm} \label{thm:LOGs_eq_EGs}
The interior of the linear ordering polytope $\co(\mathcal{H_R})$ and the set of all possible expert graphs $\mathcal{G}$ are equivalent. That is, $\reg = (\Classes, E, f_d(\cdot)) \in \mathcal{G}$ if and only if there exists $H_w = (\Classes, E, f_w(\cdot)) \in \co(\mathcal{H_R})$ with $f_w(e) = f_d(e) ~\forall~ e \in E$.
\end{thm}

\begin{figure}
\centering
\begin {tikzpicture}[-latex, auto,node distance =2cm and 2cm ,on grid ,
semithick, baseline, anchor=base]
\draw [line width=.1cm] (-4,-2) rectangle (4, 2);
\filldraw[line width=.1cm, dotted, color=blue, fill = blue!4] (-3,-2.1) rectangle (3, 1);
\filldraw[line width=.1cm, color=red, fill = violet!10] (-2.9,-2) rectangle (2.9, .9);
\node at (0, -0.5){ $\color{red} \mathcal{G} \color{black} = \color{red}\cco(\mathcal{G}_\cU)  \color{black} = \color{blue} \co(\mathcal{H_R}) \color{black} = \color{blue} \mathcal{H}$};
\node[color = blue] at (3.5, -.5) { $\mathcal{H_R}$};
\node[color = red] at (-3.5, -.5) { $\mathcal{G}_\cU$};
\node at (0, 1.4) {CCWDs};
\end{tikzpicture}
\caption{An abstract diagram showing the membership of various sets defined in this section, building on Figure~\ref{fig:sets_curl}. The set of rankings $\color{blue}\mathcal{H_R}\color{black}$ is shown as blue dots because they are discrete structures. These rankings can have cycles with curl $=1$ or $=\ell-1$, so they are shown to escape the curl condition set on the boundary.  Theorem~\ref{thm:LOGs_eq_EGs} shows that $\color{red}\mathcal{G}\color{black} = \color{blue}\co(\mathcal{H_R}) \color{black}$. Finally, the gap between the CCWDs and $\color{blue}\co(\mathcal{H_R}) \color{black}$ comes from literature on the linear ordering polytope \cite{fishburn}.}
\label{fig:sets}
\end{figure}
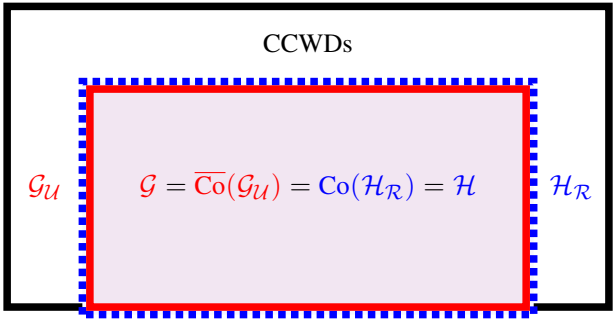

The next two subsections will focus on proving this equivalence. In Subsection~\ref{sec:LOPtoExperts}, we will give a reduction from any LOG to a probability table that generates an expert graph with the same pairwise edge weights. In Subsection~\ref{sec:ExpertstoLOP}, we will give a reduction from any expert graph to a set of rankings $r \in \mathcal{R}$ and weights $w(r)$ that generate a LOG that matches the expert graph's pairwise weights. This equivalence will allow us to harness results from the linear ordering polytope framework to understand the power and limitations of the curl condition.

\subsection{Reduction from LOGs to expert graphs}\label{sec:LOPtoExperts}
To show that any LOG can also be an expert graph, we will give a mapping from $w(\cdot): \mathcal{R} \mapsto [0,1]$ to a set of states $U$ with categorical distributions $\vec{p}_u$ and probabilities $d(\cdot): \mathcal{U} \mapsto (0, 1)$ such that $f_d(\cdot) = \E_d[f_u(\cdot)]$ matches  $f_w(\cdot) = \E_w[f_r(\cdot)]$.

We will first show how to give a set of class probabilities $\vec{p}_u$ for which the pairwise expert output gets arbitrarily close to that of a ranking $r\in R$. Here it will be convenient to think of the edge-weights as a vector $\vec{f_u} = (f_u(e_1), \ldots, f_u(e_m))$ and $\vec{f_d} = (f_d(e_1), \ldots, f_d(e_m))$.
\begin{lem}
\label{lem:almosttournaments}
Let $H_r = (\Classes, E, f_r(\cdot))$ be a ranking graph with $r = (r_1, \ldots, r_n)$. For all $\varepsilon>0$, we can construct a situational expert graph $\sit = (\Classes, E, f_u(\cdot))$ with a categorical distribution given by $p_u^{(1)},\ldots, p_u^{(n)} \in (0, 1)$ such that 
\begin{equation}
    \twonorm{\vec{f_u} - \vec{f_r}} = \sum_{e\in E} (f_r(e) - f_u(e))^2 < \varepsilon.
\end{equation}
\end{lem}
\begin{IEEEproof}
Let $t \in \R$. Let $\alpha_{i} = \frac{1}{t^{i}}$ and $z = \sum_{i=1}^n \alpha_{i}$. If we set $p_u^{(r_i)} = \frac{\alpha_{i}}{z}$, then
\begin{equation*}
    f^{(t)}_u(\YY{r_i}, \YY{r_j}) \;:\; \begin{cases}
        \leq \frac{1}{t} & \text{if } i  > j\\
        =\frac{1}{2} &  \text{if } i = j\\
        >  1 - \frac{1}{t} & \text{if } i < j\\
    \end{cases}.
\end{equation*}

Recall that $f_r(\YY{r_i}, \YY{r_j}) = \indicator{i<j}$. Thus, we have assigned successively larger probabilities to higher ranked classes such that the resulting experts have 
 \begin{equation}
     \abs{f^{(t)}_u(\YY{r_i}, \YY{r_j}) - f_r(\YY{r_i}, \YY{r_j})} \leq \frac{1}{t}.
 \end{equation}
 There are $m$ edges, so set $\frac{1}{t} = \frac{\varepsilon}{m}$ to get
$\twonorm{\vec{f_u}^{(t)} - \vec{f_r}} \leq \varepsilon$ as desired.
\end{IEEEproof}

We can use Lemma~\ref{lem:almosttournaments} to construct a state $u_r$ with probability vector $\vec{p}_u$ which achieves a $f_u(\cdot)$ that is \emph{very close} to $f_r(\cdot)$ for each $r \in \mathcal{R}$.

The remainder of this reduction is given by showing the interior of the convex hull of our generated $\vec{f_u}$ vectors is the same as the interior of the convex hull of $\vec{f_r}$ vectors. To do this, we make use of the following more general result:

\begin{thm}
\label{thm:convex_hulls}
Consider a set $V = \{\vec{v_1}, \ldots, \vec{v_t} \}$ and with $\vec{v_i} \in \R^m$ for all $i$. If we have $\Vt$ such that for every $\varepsilon$ and $\vec{v} \in V$, there exists $\vec{\vt} \in \Vt$ such that $\twonorm{\vec{\vt} - \vec{v}} < \varepsilon$, then we have
\begin{equation}
    \co(V) \subseteq \cco(\Vt).
\end{equation}
\end{thm}
\begin{IEEEproof}
    The proof is the result of Lemma~\ref{lem:convex_analysis}, which we prove in Appendix Section~\ref{apx:convex}.
\end{IEEEproof}

This machinery allows us to prove the goal lemma of this subsection:
\begin{lem}\label{lem:rsubsetG}
$\co(\mathcal{H_R}) = \mathcal{G}.$
\end{lem}
\begin{IEEEproof}
Consider Theorem~\ref{thm:convex_hulls}. Let $V = \{\vec{f_r} \::\: H_r \in \mathcal{H_R}\}$ and $\Vt =\{ \vec{f_u} \::\: (\Classes, E, f_d(\cdot))\in \mathcal{G}_\cU$. Lemma~\ref{lem:almosttournaments} shows that for every $\varepsilon$ and $H_r \in \mathcal{H_R}$ there is some $\sit \in \mathcal{G}_\cU$ for which the edge-weight vectors are within $\varepsilon$:  $\twonorm{\vec{f_u} - \vec{f_r}} < \varepsilon$. This satisfies the requirements for us to apply Theorem~\ref{thm:convex_hulls} to say $\co(\mathcal{H_R}) = \cco(\mathcal{G}_\cU) = \mathcal{G}$.
\end{IEEEproof}

\subsection{Reduction from expert graphs to LOGs} \label{sec:ExpertstoLOP}
\begin{figure*}
\centering
\begin{align*}
    \underbrace{\left(\scalebox{.7}{\begin {tikzpicture}[-latex, auto,node distance =1cm and 1.33 cm ,on grid,
semithick, baseline, anchor=base, state/.style ={ circle, draw, minimum width =.8 cm}]
\node (center) {};
\node[state] (c) [below left = of center]{$.2$};
\node (clab) [below left = .5cm and .5cm of c]{$\YY{3}$};
\node[state] (a) [above = of center]{$.5$};
\node (alab) [above = .7cm of a]{$\YY{1}$};
\node[state] (b) [below right = of center]{$.3$};
\node (blab) [below right = .5cm and .5cm of b]{$\YY{2}$};
\path (a) edge node[above right] {$\frac{5}{8}$} (b);
\path (b) edge node[below] {$\frac{3}{5}$} (c);
\path (c) edge node[above left] {$\frac{2}{7}$} (a);
\end{tikzpicture}}\right)}_{\sit} &= .5 \underbrace{\left(\scalebox{.7}{\begin {tikzpicture}[-latex, auto,node distance =1cm and 1.33 cm ,on grid,
semithick, baseline, anchor=base, state/.style ={ circle, draw, minimum width =.8 cm}]
\node (center) {};
\filldraw[color = red, fill=red!4](-2.3, -.5) rectangle (2.3, -1.8); \node[color = red] at (0,-.3) {$G^{\vec{p}[-1]}_u$};
\node[state, color=red] (c) [below left = of center]{$.4$};
\node (clab) [below left = .5cm and .5cm of c]{$\YY{3}$};
\node[state, fill] (a) [above = of center]{$.5$};
\node (alab) [above = .7cm of a]{$\YY{1}$};
\node[state, color=red] (b) [below right = of center]{$.6$};
\node (blab) [below right = .5cm and .5cm of b]{$\YY{2}$};
\path (a) edge node[above right] {$1$} (b);
\path (b) edge[color=red] node[below] {$\frac{3}{5}$} (c);
\path (c) edge node[above left] {$0$} (a);
\end{tikzpicture}}\right)}_{\sit[1]} + .3 \underbrace{\left(\scalebox{.7}{\begin {tikzpicture}[-latex, auto,node distance =1cm and 1.33 cm ,on grid,
semithick, baseline, anchor=base, state/.style ={ circle, draw, minimum width =.8 cm}]
\node (center) {};
\node[state, color=red] (c) [below left = of center]{$\frac{2}{7}$};
\node (clab) [below left = .5cm and .5cm of c]{$\YY{3}$};
\node[state, color=red] (a) [above = of center]{$\frac{5}{7}$};
\node (alab) [above = .7cm of a]{$\YY{1}$};
\node[state, fill] (b) [below right = of center]{$.6$};
\node (blab) [below right = .5cm and .5cm of b]{$\YY{2}$};
\path (a) edge node[above right] {$0$} (b);
\path (b) edge node[below] {$1$} (c);
\path (c) edge[color=red] node[above left] {$\frac{2}{7}$} (a);
\begin{pgfonlayer}{bg}
\filldraw[color = red, fill=red!4]($(a) + (-.8, 0) + .5*(a)-.5*(c)$) --($(a) + (.7, 0) + .5*(a)-.5*(c)$) -- ($(c) + (.7, 0)- .4*(a)+.4*(c)$) -- ($(c) + (-.8, 0) - .4*(a)+.4*(c)$) -- cycle;
\end{pgfonlayer}
\node[color = red] at ( $ (a)!0.5!(c) + (-1, .5)$ ) {$G^{\vec{p}[-2]}_u$};
\end{tikzpicture}}\right)}_{\sit[2]} + 
.2 \underbrace{\left(\scalebox{.7}{\begin {tikzpicture}[-latex, auto,node distance =1cm and 1.33 cm ,on grid,
semithick, baseline, anchor=base, state/.style ={ circle, draw, minimum width =.7 cm}]
\node (center) {};
\node[state, fill] (c) [below left = of center]{$.4$};
\node (clab) [below left = .5cm and .5cm of c]{$\YY{3}$};
\node[state, color=red] (a) [above = of center]{$\frac{5}{8}$};
\node (alab) [above = .7cm of a]{$\YY{1}$};
\node[state, color=red] (b) [below right = of center]{$\frac{3}{8}$};
\node (blab) [below right = .5cm and .5cm of b]{$\YY{2}$};
\path (a) edge[color=red] node[above right] {$\frac{5}{8}$} (b);
\path (b) edge node[below] {$0$} (c);
\path (c) edge node[above left] {$1$} (a);
\begin{pgfonlayer}{bg}
\filldraw[color = red, fill=red!4]($(a) + (-.7, 0) + .5*(a)-.5*(b)$) --($(a) + (.8, 0) + .5*(a)-.5*(b)$) -- ($(b) + (.8, 0)- .4*(a)+.4*(b)$) -- ($(b) + (-.7, 0) - .4*(a)+.4*(b)$) -- cycle;
\end{pgfonlayer}
\node[color = red] at ( $ (a)!0.5!(b) + (1, .5)$ ) {$G^{\vec{p}[-3]}_u$};
\end{tikzpicture}}\right)}_{\sit[3]}
\end{align*}
\caption{A decomposition of an expert graph $\sit$ into prefix expert graphs $\sit[1], \sit[2], \sit[3]$ following Lemma~\ref{lem:situational_EG_to_prefixGraph}. Probabilities from probability table $\vec{p}$ are shown inside the nodes in $\sit$ and probabilities for $\vec{p}[-i]$ are shown in red inside nodes in $G^{\vec{p}[-i]}_u$. To get a decomposition into rankings $H_r$ we apply the same procedure to the highlighted sub-graphs $G^{\vec{p}[-1]}_u, G^{\vec{p}[-2]}_u, G^{\vec{p}[-3]}_u$. This determines the second choice distribution for each of these sub-populations, which will imply a third choice as well, giving a decomposition into rankings.}
\label{fig:eg_to_lop_decomp}
\end{figure*}
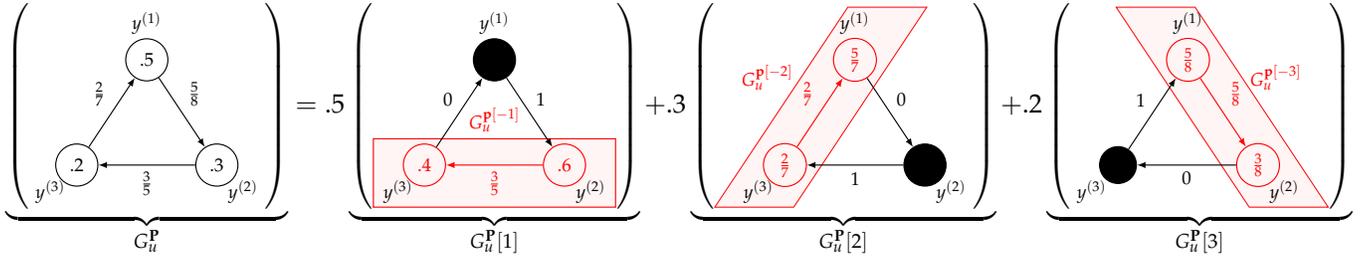

We will now show that, given an expert graph $\reg = (\mathcal{C}, E, f_d(\cdot))$, we can find a decomposition into ranking graphs. Following the usual strategy in this paper, we will begin by showing we can decompose a situational expert graph $\sit$ into LOGs, which will imply the overall result. Recall that any true expert graph \emph{must} be generated by some probability table. Lemma~\ref{lem:situational_EG_to_prefixGraph} will sort the population into sub-populations for each \emph{first} choice and determine their sizes from values in the probability table. A recursion on each sub-population will then determine the distribution of second choices, and so on. A sketch of this procedure is given in Figure~\ref{fig:eg_to_lop_decomp}.

In order to describe the expert graph on the sub-population with a set of choices, we will define ``prefix expert graphs.''
\begin{defn}
Given an expert graph $\reg=(\Classes, E, f_d(\cdot))$ and prefix $a_1$ and non-specified leftover indices $\{b_1, \ldots b_{n-1}\}$, we define the \textbf{prefix expert graph} $\reg[a_1] = (\Classes, E, f_{\reg[a_1]}(\cdot))$ to be a graph with edge weights given by:
\begin{align*}
    f_{\reg[a]}(\YY{a_1}, \YY{b_j}) &= 1, \\
    f_{\reg[a]}(\YY{b_i}, \YY{b_j}) &= f_{d}(\YY{b_i}, \YY{b_j}).
\end{align*}
\end{defn}
The prefix graph fixes a primary preference on $C^{(a_1)}$ while maintaining relationships between $C^{(b_1)}, \ldots C^{(b_{n-1})}$.

\begin{lem}\label{lem:situational_EG_to_prefixGraph}
We can decompose any situational expert graph $\sit = (\Classes, E, f_u(\cdot))$ with categorical distribution $\vec{p}_u = (p_u^{(1)}, p_u^{(2)}, \ldots, p_u^{(n)}) \in \simplex^{n}$ into prefix expert graphs with a distribution given by
\begin{equation*}
    \sit = \sum_{a_1=1}^n p_u^{(a_1)} \sit[a_1].
\end{equation*}
That is, the edge weights
\begin{equation*}
    f^*(e) = \sum_{a_1=1}^n p_u^{(a_1)} f_{\sit[a_1]}(e).
\end{equation*}
\end{lem}
\begin{IEEEproof}
For this decomposition, the desired edge-weights $f^*(e)$ for $e = (\YY{i}, \YY{j})$ are achieved:
\begin{align*}
    f^*(e) &= p_u^{(i)}\underbrace{f_{\sit[i]}(e)}_{=1} + p_u^{(j)} \underbrace{f_{\sit[j]}(e)}_{=0} + \sum_{a_1 \neq i,j} p_u^{(a_1)} \underbrace{f_{\reg[a_1]}(e)}_{=f_u(e)}\\
    &= p_u^{(i)} + \sum_{a_1 \neq i,j} p_u^{(a_1)} f_u(e)\\
     &= \left(p_u^{(i)} + p_u^{(j)}\right) \underbrace{\frac{p_u^{(i)}}{p_u^{(i)} + p_u^{(j)}}}_{f_u(e)}+ \sum_{a_1 \neq i,j} p_u^{(a_1)} f_u(e)\\
    &= \sum_{a_1 = 1}^n p_u^{(b_1)} f_u(e) = f_u(e)\\
\end{align*}
\end{IEEEproof}

After Lemma~\ref{lem:situational_EG_to_prefixGraph} decides on the sub-population distribution for first choice candidates, we will recurse on each sub-population to determine the distribution on their second choice candidates.
\begin{defn}
We define $G^{\vec{p}[-i]}_u$ to be the the subgraph of $G^{\vec{p}}_u$ with $\YY{i}$ removed:
\begin{equation*}
    G^{\vec{p}[-i]}_u = (\Classes \setminus \{\YY{i}\}, E \setminus \{e \;:\; \YY{i} \in e\}, f^{(-i)}_u(\cdot)).
\end{equation*}
\end{defn}

\begin{obs}\label{obs:newprobtable}
The subgraph $G^{\vec{p}[-i]}_u$ is generated by a new probability table with 
\begin{equation}
    p[-i]_u^{(j)} = \frac{p_u^{(j)}}{1 - p_u^{(i)}}.
\end{equation}
\end{obs}

\begin{lem}\label{lem:situational_EG_to_ranking}
Given a situational expert graph $\sit = (\Classes, E, f_u(\cdot))$ with categorical distribution $\vec{p}_u = (p_u^{(1)}, p_u^{(2)}, \ldots, p_u^{(n)}) \in \simplex^{n}$,
\begin{equation}\label{eq:rankingweights}
    w(r) = \prod_{i=1}^{n} \frac{p_u^{(r_i)}}{\prod_{j=1}^{i-1} (1 - p_u^{(r_j)})} = \frac{\prod_{i=1}^{n} p_u^{(r_i)}}{\prod_{i=0}^{n-1} (1 - p_u^{(r_i)})^{n-i}}
\end{equation}
generates $H_w = (\Classes, E, f_w(\cdot))$ with $f_w(e) = f_u(e) ~\forall~ e \in E$.
\end{lem}
\begin{IEEEproof}
Lemma~\ref{lem:situational_EG_to_prefixGraph} gives the decomposition
\begin{equation*}
    \sit = \sum_{i=1}^n p_u^{(i)} \sit[i].
\end{equation*}
Now, since the behavior of all edges with $\YY{i} \in e$ is the same for all components in $\sit[i]$, we can recurse on the sub-graph with $\YY{i}$ removed, with probabilities defined by Observation~\ref{obs:newprobtable}. The base case of $\Classes= \{\YY{j}\}$ is trivial, with $p_u^{(j)} = 1$. Multiplication of the probabilities adjusted by Observation~\ref{obs:newprobtable} gives Equation~\ref{eq:rankingweights}.
\end{IEEEproof}

Finally, we can give the desired result for this section.
\begin{lem}\label{lem:gsubsetr}
$\mathcal{G} \subseteq \co(\mathcal{R}).$
\end{lem}
\begin{IEEEproof}
Pick $\reg \in \mathcal{G}$ without loss of generality. Decompose each state $u$ using probability vector $\vec{p}_u$ according to Lemma~\ref{lem:situational_EG_to_ranking}. Recall from Assumption~\ref{assum:0} that all class probabilities $p_u^{(i)}>0$, which yields $w(r)>0 ~\forall~ r \in \mathcal{R}$. Hence, $\reg \subseteq \co(\mathcal{R})$.
\end{IEEEproof}

\subsection{Implications for sufficiency of the curl condition}
Theorem~\ref{thm:convex_hulls} allows us to harness results from the voting theory to show our condition is sufficient to describe all expert graphs with $n \leq 5$ classes \cite{fishburn}. The curl condition is insufficient for the linear ordering polytope with $n \geq 6$.

Theorem~\ref{thm:convex_hulls} can also be used as a proof technique for sufficiency on more specific classes of expert graphs. The task now reduces to finding a decomposition of the weighted directed graph into a convex combination of rankings, for which we can find class probabilities via Lemma~\ref{lem:almosttournaments}. Figure \ref{fig:situational_LOG_decomp} gives an example of such a decomposition of a cycle. Importantly, these decompositions cannot depend on a probability table $\vec{P}$ or distribution $d(\cdot)$ as we did in Subsection~\ref{sec:ExpertstoLOP}, since the goal here is to show that such a $\vec{P}$ and distribution $d(\cdot)$ \emph{exist}.

We can show sufficiency of the curl condition for expert \emph{cycles} via a decomposition algorithm of any curl-consistent expert cycle into ranking graphs that generate the edge-weights. The existence of such a decomposition is shown in Appendix Section~\ref{apx:sufficiencyexistence}. A constructive decomposition is also given Appendix Section~\ref{apx:decomposecycles}, which is demonstrated in Figure~\ref{fig:knowledge_graph_decomp_algo}.

\section{Conclusion}\label{sec:conclude}
We have defined expert graphs as a framework to understand the amalgamation of pairwise experts with overlapping training. To analyze these graphs, we introduced the curl and derived \emph{necessary} lower and upper bounds. From this notion of curl we provided an algorithm to derive upper and lower bounds on missing edges of an expert graph. We have also shown our expert graph framework's characterization to be equivalent to that of the linear ordering polytope, which allows us to apply results from a rich literature \cite{Noga, fishburn, McGarvey}.

Future work may expand the notion of the expert graph to an expert \emph{hypergraph}, which includes experts with potentially \emph{nonbinary} domains of expertise.

One of the most relevant applications of this framework is in machine learning, where a practitioner may use a number of classifiers trained on different overlapping datasets. After being given additional information, the user may wish to create a ``synthetic classifier'' for a new prior on the distribution of labels. Hence, our results point towards an exciting new way of handling label shift after training.

\newpage
\appendix

\subsection{Proof of Theorem~\ref{thm:convex_hulls}}
\label{apx:convex}
Convex hulls of finite sets in $\R^\l$ are \emph{convex} polytopes, which can be expressed as an intersection of $h$ halfspaces indexed by $f$ with $\{\vec{x} : \vec{a^{(f)}}^\top \vec{x} < b^{(f)}\}$ \cite{polytopes}. The perpendicular vectors $\vec{a^{(f)}}^\top$ can be combined as row-vectors of the matrix $A$ so that any convex polytope can be expressed as
\begin{equation}
\label{eq:polytopes}
 \{x : A\vec{x} \prec \vec{b} \} = \left\{\vec{x} : \begin{pmatrix} (\vec{a^{(1)}})^\top\\ \vdots \\ (\vec{a^{(h)}})^\top\end{pmatrix}\vec{x} \prec \begin{pmatrix}b^{(1)}\\ \vdots \\ b^{(h)} \end{pmatrix} \right\}
\end{equation}
For convenience, the vectors $\vec{a^{(f)}}, \vec{\at^{(f)}}$ are assumed to be unit vectors throughout.

\begin{lem}
\label{lem:convex_analysis}
Given $V = \{v_1, \ldots, v_m \}$ and ``perturbed points'' $\Vt = \{\vt_1, \ldots \vt_m \}$ with $v_i \in \R^\l$ and $\vt_j \in \R^
\l$ for all $i, j$. We have that if $\vec{x} \in \co(V)$ and is $\varepsilon >0$ from the boundary $\bo(\co(V))$, then if we can find perturbed points $\Vt$ such that they are within $\varepsilon$ from the desired $V$, then $\vec{x} \in \cco(\Vt)$.

More precisely, let
\begin{align*}
    \co(V) &= \{\vec{x} : A\vec{x} \prec \vec{b}\}\\
    \co(\Vt) &= \{\vec{x} : \At \vec{x} \prec \vec{\bt}\}
\end{align*}
as given by Equation~\ref{eq:polytopes}. If $A \vec{x} \prec \vec{b} - \varepsilon \onevec$ and $\twonorm{\vec{v_i} - \vec{\vt_i}} < \varepsilon\; \forall i$, then $\At \vec{x} \prec \vec{\bt}$.
\end{lem}

To prove this theorem, we will need to show that the boundaries of the polytopes do not move too far. We will do this using Lemma~\ref{lem:perturbed_boundaries}, which bounds how far $\bo(\co(V))$ can be from $\bo(\co(\Vt))$ along a single ``face.'' 

\begin{defn}
\label{defn:arbitrary_face}
Choose $f \in [h]$. Define:
\begin{align*}
    W^{(f)} &= \{\vec{w} : (\vec{a^{(f)}})^\top w = b^{(f)}, \vec{w} \in V\}\\
    \Wt^{(f)} &= \{\vec{\vt_i} : \vec{v_i} \in W^{(f)}\}
\end{align*}
We restrict the size of $\abs{W^{(f)}} = \l$, which is the number of points needed to define a halfspace in $\R^\l$. This can be done by allowing for multiple identical $\vec{a_f}, b_f$ combinations corresponding to all size $l$ subsets of the $v_i$ along the boundary. 

Note that $\co(W^{(f)})$ describes a ``face'' of the polytope $\co(V)$ indexed by $f$ which is perpendicular to $\vec{a^{(f)}}$. $\co(\Wt^{(f)})$ describes the perturbed face.
\end{defn}

\begin{lem}
\label{lem:perturbed_boundaries}
Choose $f, g \in [h]$ arbitrarily and let $W^{(f)} = \{\vec{w^{(f)}_1}, \ldots, \vec{w^{(f)}_\l}\}$ and $\Wt^{(f)} = \{\vec{\wt^{(f)}_1}, \ldots, \vec{\wt^{(f)}_\l}\}$.  
For every $\vec{m^{(f)}} \in \cco(W^{(f)})$, we have $(\vec{\at^{(g)}})^\top \vec{m^{(f)}} < \bt^{(g)} + \varepsilon$.
\end{lem}
\begin{IEEEproof}
Because $m \in \cco(W^{(f)})$, there is some $\lambda \in \triangle_\l$ with
\begin{equation}
    \vec{m^{(f)}} = \sum_{i=1}^{\l} \lambda_i \vec{w^{(f)}_i} \in \cco(W^{(f)})
\end{equation}
Consider also
\begin{equation}
    \vec{\mt^{(f)}} = \sum_{i=1}^{\l} \lambda_i \vec{\wt^{(f)}_i} \in \cco(\Wt^{(f)})
\end{equation}
Note that the norm of the difference between these two vectors is bounded:
\begin{equation}
\begin{aligned}
    \twonorm{\vec{m^{(f)}} - \vec{\mt^{(f)}}} &= \twonorm{\sum_{i=1}^{\l} \lambda_i (\vec{w^{(f)}_i} - \vec{\wt^{(f)}_i})} \\
    &\leq \sum_{i=1}^{\l} \lambda_i \underbrace{\twonorm{\vec{w^{(f)}_i} - \vec{\wt^{(f)}_i}}}_{< \varepsilon} < \varepsilon
\end{aligned}
\end{equation}
Also note that because $\vec{\mt^{(f)}}\in \cco(\Wt^{(f)}) \subseteq \cco(\Vt)$, we have that $(\vec{\at^{(g)}})^\top \vec{\mt^{(f)}} \leq \bt^{(g)}$. Now, a simple application of Cauchy-Schwartz gives:
\begin{equation}
    \begin{aligned}
        (\vec{\at^{(g)}})^\top \vec{m^{(f)}} &= (\vec{\at^{(g)}})^\top(\vec{\mt^{(f)}} + (\vec{m^{(f)}} - \vec{\mt^{(f)}}))\\
        &= \underbrace{(\vec{\at^{(g)}})^\top \vec{\mt^{(f)}}}_{\leq \bt^{(g)}} + (\vec{\at^{(g)}})^\top (\vec{m^{(f)}} - \vec{\mt^{(f)}})\\
        &\leq \bt^{(g)} + \twonorm{\vec{\at^{(g)}}}\twonorm{\vec{m^{(f)}} - \vec{\mt^{(f)}}}\\
        &< \bt^{(g)} + \varepsilon
    \end{aligned}
\end{equation}
\end{IEEEproof}

With this, we are now ready to prove Lemma~\ref{lem:convex_analysis}.
\begin{IEEEproof}
Choose an arbitrary face $g \in [h]$. Recall we have $\vec{x} \in \co(V)$ with $(\vec{a^{(g)}})^\top \vec{x} < b - \varepsilon$ and we wish to show $(\vec{\at^{(g)}})^\top \vec{x} < \bt^{(g)}$.

Let $\vec{m^{(f)}_x}$ be the result of extending $\vec{\at^{(g)}}$ from $\vec{x}$ to $\bo(V)$. This must hit some face with $(\vec{a^{(f)}})^\top \vec{m^{(f)}_x} = b^{(f)}$, so $\vec{m^{(f)}_x} \in \co(W^{(f)})$. That is, find $\beta$ such that
\begin{equation}
        \vec{m^{(f)}_x} = \beta \vec{\at^{(g)}} + \vec{x} \in \co(W^{(f)})
\end{equation}

First, lets bound $\beta$. Notice that because $\vec{m^{(f)}_x} \in \co(W^{(f)})$, we have
\begin{equation}
\begin{aligned}
    (\vec{a^{(f)}})^\top\vec{m^{(f)}_x} &= (\vec{a^{(f)}})^\top \left( \sum_{i=1}^{\l} \lambda_i \vec{w^{(f)}_i}\right) \\
    &= \sum_{i=1}^{\l} \lambda_i(\vec{a^{(f)}})^\top \vec{w^{(f)}_i} = b^{(f)}
\end{aligned}
\end{equation}
So, we have
\begin{equation}
    b^{(f)} = (\vec{a^{(f)}})^\top\vec{m^{(f)}_x} = \beta \underbrace{(\vec{a^{(f)}})^\top \vec{\at^{(g)}}}_{\leq 1} + \underbrace{(\vec{a^{(f)}})^\top \vec{x}}_{< b^{(f)} - \varepsilon} \Rightarrow \varepsilon < \beta
\end{equation}
Now, apply Lemma~\ref{lem:perturbed_boundaries}
\begin{equation}
    \begin{aligned}
        (\vec{\at^{(g)}})^\top \vec{m^{(f)}_x} &< \bt^{(g)} + \varepsilon\\
        (\vec{\at^{(g)}})^\top \vec{x} + (\vec{\at^{(g)}})^\top \vec{\at^{(g)}} \beta &< \bt^{(g)} + \varepsilon\\
        (\vec{\at^{(g)}})^\top \vec{x} &< \bt^{(g)}
    \end{aligned}
\end{equation}
Recall we chose face $g \in [h]$ arbitrarily, so this holds for all halfspaces in the convex polytope. Hence, we have $A \vec{x} \prec \vec{b}$.
\end{IEEEproof}

\subsection{Existence proof for decomposing expert cycles into LOGs}\label{apx:sufficiencyexistence}
We can show that a decomposition of any curl consistent cycle into acyclic orientations must exist. Consider $\vec{v}$ the vector of edge-weights of an arbitrary cycle with $v^{(i)} = f_d^{(\YY{i}, \YY{i+1})}$. Acyclic orientations in this framework correspond to the set $A = \{0, 1\}^\l \setminus \{\vec{0}_\ell, \vec{1}_\l\}$.

\begin{lem}\label{lem:decomp_existance}
 Given $\vec{v}\in (0,1)^{\ell}$, such that $1< \|\vec{v}\|_1<\ell-1$, then for some $d(\cdot): A \mapsto (0, 1)$ in  $\triangle_{2^\l - 2}$, we have
\begin{equation}
\vec{v} = \sum_{\vec{t} \in G} d(\vec{t}) \vec{t}
\end{equation}
\end{lem}
\begin{IEEEproof}
Consider $H = \{\vec{h}\in [0,1]^{\ell}: 1\leq \|\vec{h}\|_1 \leq \ell-1\}.$ Note that $\vec{v} \in H$.  
Since $A$ and $H$ are closed and bounded sets, they are compact. Further, $H$ is convex. Therefore, $\co(A)$ is also convex and compact using compactness and Carathéodory's theorem~\cite{Bertsimas}.
We can now prove the required statement by showing $H = \cco(A)$. 

\begin{enumerate}
\item\emph{$\cco(A)\subset H$:} Let $\vec{y} \in \cco(A)$, then $\vec{y} = \sum_{\vec{t}\in A} d(\vec{t})\vec{t}$, for some $d(\cdot): A \mapsto (0, 1)$ in  $\triangle_{2^\l - 2}.$ We notice that $\vec{y} \in [0,1]^{\ell}$ and $\|\vec{y}\|_1  \in [1,\ell-1].$ Therefore $\vec{y}\in H$ and hence $\cco(A)\subset H.$
\item \emph{$H\subset \cco(A)$:} Let $E$ be the set of extreme points of $H$. Since $H$ is convex and compact, we can use the Krein-Milman Theorem~\cite{Bertsimas} to get $H = \cco(E)$. Further, we notice that $E\subset A$~\cite{Bertsimas}, therefore $H\subset \cco(A).$
\end{enumerate}
\end{IEEEproof}

\subsection{Algorithmically decomposing expert cycles into LOGS}
\label{apx:decomposecycles}
For this section, consider a knowledge cycle $G = (\Classes, E, f(\cdot))$ with edges $\YY{1} \rightarrow \YY{2} \rightarrow \ldots \rightarrow \YY{n} \rightarrow \YY{1}$. Let the edge weights be represented by a vector $\vec{f_0} \in (0, 1)^{n}$ with $f_0^{(i)} = f_d^{(\YY{i}, \YY{i+1})}$ for $i < n$, and $f_0^{(n)} = f_d^{(\YY{n}, \YY{1})}$. We will refer to \emph{not fully decomposed} components as vectors $\vec{f_j} \in [0, 1]^\l$ and all \emph{orientation} components as $\vec{t_j} \in \{0, 1\}^{\l}$ with weight $d(\vec{t_j})$. We will also use $\indvec{I} \in \{0, 1\}^\l$ to denote a vector for which $\indvec{I}^{(i)} = \indicator{i \in I}$.

We begin by observing that all $t_j\in\{0, 1\}^\l$ are allowed other than $\onevec$ and the origin. 
\begin{obs}\label{obs:decompose_cycle}
We can decompose a scaled version of $\onevec$ as follows:
\begin{equation}
    \left(1 - \frac{1}{\l}\right)\onevec = \sum_{i=1}^{\l-1} \frac{ \indvec{\{j: j \neq i\}}}{\l} = \sum_{i=1}^{\l-1} \frac{ \indvec{-\{j\}}}{\l}
\end{equation}
\end{obs}
\begin{obs}\label{obs:finishing_decomp}
If $\onenorm{\vec{f_j}} = 1$, we can decompose:
\begin{equation}
    \vec{f_j} = \sum_{i=1}^{\l-1} f_j^{(i)} \indvec{\{i\}}
\end{equation}
\end{obs}

\newcommand{\supp}{\text{Supp}}

\begin{defn}
We define the support of vector $\vec{f_j} \in [0, 1]^\l$ to be the set of nonzero indices.
\begin{equation}
    \supp(\vec{f_j}) = \{i : \vec{f_j}^{(i)} > 0]\}
\end{equation}
\end{defn}

\begin{lem}
\label{lem:decrease_support}
Let $\gamma$ be a factor for handling the case of full support:
\begin{equation}
    \gamma = 1 - \frac{\indicator{\abs{S_j} = \l}}{\l}
\end{equation}
Given $\vec{f_j}\in[0,1]^\l$ with $\supp(\vec{f_j}) = S_j$ and $\onenorm{\vec{f_j}} \in (1, \gamma \abs{S_j})$, we can decompose
\begin{equation}
    \vec{f_j} = z_j \gamma \indvec{S_j} + (1 - z_j) \vec{f_{j+1}}
\end{equation}
where either $\onenorm{\vec{f_{j+1}}} = 1$ or
\begin{enumerate}[(i)]
    \item $\abs{\supp(\vec{f_{j+1})}} \leq \abs{S_j} - 1$.
    \item $\onenorm{\vec{f_{j+1}}} \in (1, \abs{S_j} - 1)$.
\end{enumerate}
\end{lem}
\begin{IEEEproof}
We begin by first observing
\begin{equation}
    f_{j+1}^{(i)} = \frac{f_{j}^{(i)} - z_j \gamma \indicator{i \in S_j}}{1 - z_i} \leq \frac{f_{j}^{(i)} - z_j \indicator{i \in S_j}}{1 - z_i}
\end{equation}
Recall that if $f_{j}^{(i)} > 0$, then $i \in S_j$. Hence $\vec{f_{j+1}} \in [0, 1]^{\l}$. We notice that the upper bound of (ii) $\onenorm{\vec{f_{j+1}}} \leq \abs{S_j} - 1$ now follows from (i). It remains to find $z_j$ such that $\onenorm{\vec{f_{j+1}}} \geq 1$ and if $\onenorm{\vec{f_{j+1}}} > 1$ then (i) holds.
 \begin{equation}
 \begin{aligned}
     \zeta_a &= \frac{1}{\gamma}\min_{i\in S_j} {f_j^{(i)}}\\
     \zeta_b &= \frac{\onenorm{\vec{f_j}} - 1}{\gamma\abs{S_j} - 1}\\
     z_j &= \min(\zeta_a, \zeta_b)
 \end{aligned}
\end{equation}
We can apply the Pigeonhole principle to $\sum_{i \in S_j} f_j^{(i)} \leq \l - 1$ to get $z_j \leq \zeta_a \leq 1$. If $z_j = 1$ then the second term of the decomposition is irrelevant. We can treat the L1 norm as a linear function when the vectors are all the same sign giving:
\begin{equation}\label{eq:onenorms}
    \onenorm{\vec{f_{j+1}}} = \frac{\onenorm{\vec{f_j}} - z_j \gamma \abs{S_j} }{1 - z_j}
\end{equation}

Now consider two cases:
\begin{enumerate}
    \item ($z_j = \zeta_b$) Notice $\zeta_b$ has been carefully chosen so that,
    \begin{equation}
        \begin{aligned}
        \onenorm{\vec{f_{j+1}}} &= \frac{\onenorm{\vec{f_j}} - \zeta_b \gamma \abs{S_j}}{1-\zeta_b} = 1.
        \end{aligned}
    \end{equation}
    \item ($z_j = \zeta_a$) Notice that if $f_j^{(i^*)} = \min_{i\in S_j} {f_j^{(i)}}$, then $f_{j+1}^{(i^*)} = 0$, which satisfies condition (i). 
    
    To show the lower bound of (ii), rewrite Equation~\ref{eq:onenorms}:
    \begin{equation}\label{eq:onenorms2}
        \onenorm{\vec{f_{j+1}}} = \onenorm{\vec{f_j}} \frac{1 - z_j \left(\frac{\gamma \abs{S_j}}{\onenorm{\vec{f_j}}}\right)}{1 - z_j}
    \end{equation}
    Recall that $\onenorm{\vec{f_j}} < \gamma \abs{S_k}$, so $\frac{\gamma \abs{S_j}}{\onenorm{\vec{f_j}}} > 1$. So, if Equation~\ref{eq:onenorms2} gives $\onenorm{\vec{f_{j+1}}} = 1$ with $z_j = \zeta_b$, then $z_j \leq \zeta_b$ gives $\onenorm{\vec{f_{j+1}}} \geq 1$.
\end{enumerate}

\end{IEEEproof}

\begin{lem}
    We can decompose any expert cycle $\reg = (\Classes, E, f_d(\cdot))$ into a linear combination of elements of $\mathcal{H_R}$:
    \begin{equation*}
        \reg = \sum_{r \in \mathcal{R}} w(r) H_r
    \end{equation*}
    As usual, this means $f_d(e) = \sum_{r \in \mathcal{R}} w(r)f_r(e)~\forall~e\in E$.
\end{lem}
\begin{IEEEproof}
Recall that edge-weight vectors for $H_r$ can be all $\vec{t_j}\in\{0, 1\}^\l$ other than $\onevec$ and the origin. The task is now to decompose the edge-weight vector $\vec{f_0} = (f_d(\YY{1}, \YY{2}), f_d(\YY{2}, \YY{3}), \ldots f_d(\YY{\l}, \YY{1}))^\top$ into a convex combination of $\vec{t_j}$. Curl consistency gives $\onenorm{\vec{f_0}} \in (1, \l-1)$.
 
Use Lemma~\ref{lem:decrease_support} to decompose
\begin{equation}
    \vec{f_0} = z_0 \gamma \indvec{S_0} + (1 - z_0) \vec{f_{1}}
\end{equation}
If $\vec{f_0}$ had full support, then $\gamma = 1 - \frac{1}{\l-1}$ and $\indvec{S_0} = \onevec$. Thus, we can use Observation~\ref{obs:decompose_cycle} to decompose $\gamma \onevec$ into acyclic orientations. If $\supp(\vec{f_0}) < \l$, then $\gamma = 1$ and $\indvec{S_0}$ already represents an acyclic orientation.

Repeatedly apply Lemma~\ref{lem:decrease_support}. In this process if we reach some $\vec{f_j}$ with $\onenorm{\vec{f_j}} = 1$, we can apply Observation~\ref{obs:finishing_decomp} to decompose $\vec{f_j}$ into acyclic orientations. If we do not, we will terminate at $\vec{f_\l}$ with $\abs{S_\l} = 0$. See Figure~\ref{fig:knowledge_graph_decomp_algo} for an example of this decomposition.
\newcommand{\fourgraph}[5]{\scalebox{.65}{\begin {tikzpicture}[-latex, auto,node distance =1.0cm and 1.0cm ,on grid ,
semithick, baseline, anchor=base]
\node (e) {$#5$};
\node[circle,fill,inner sep=0.05pt,label=above left:$C_0$] (a) [above left = of e]{$C_0$};
\node[circle,fill,inner sep=0.05pt,label=above right:$C_1$] (b) [above right = of e]{$C_1$};
\node[circle,fill,inner sep=0.05pt,label=below right:$C_2$] (c) [below right = of e]{$C_1$};
\node[circle,fill,inner sep=0.05pt,label=below left:$C_3$] (d) [below left = of e]{$C_3$};
\path (a) edge [bend right = 0] node[above] {$#1$} (b);
\path (b) edge [bend right = 0] node[right] {$#2$} (c);
\path (c) edge [bend right = 0] node[below] {$#3$} (d);
\path (d) edge [bend right = 0] node[left] {$#4$} (a);
\end{tikzpicture}}}
\newcommand{\indvecc}[1]{\vec{1_{4}}[#1]}

\begin{figure*}
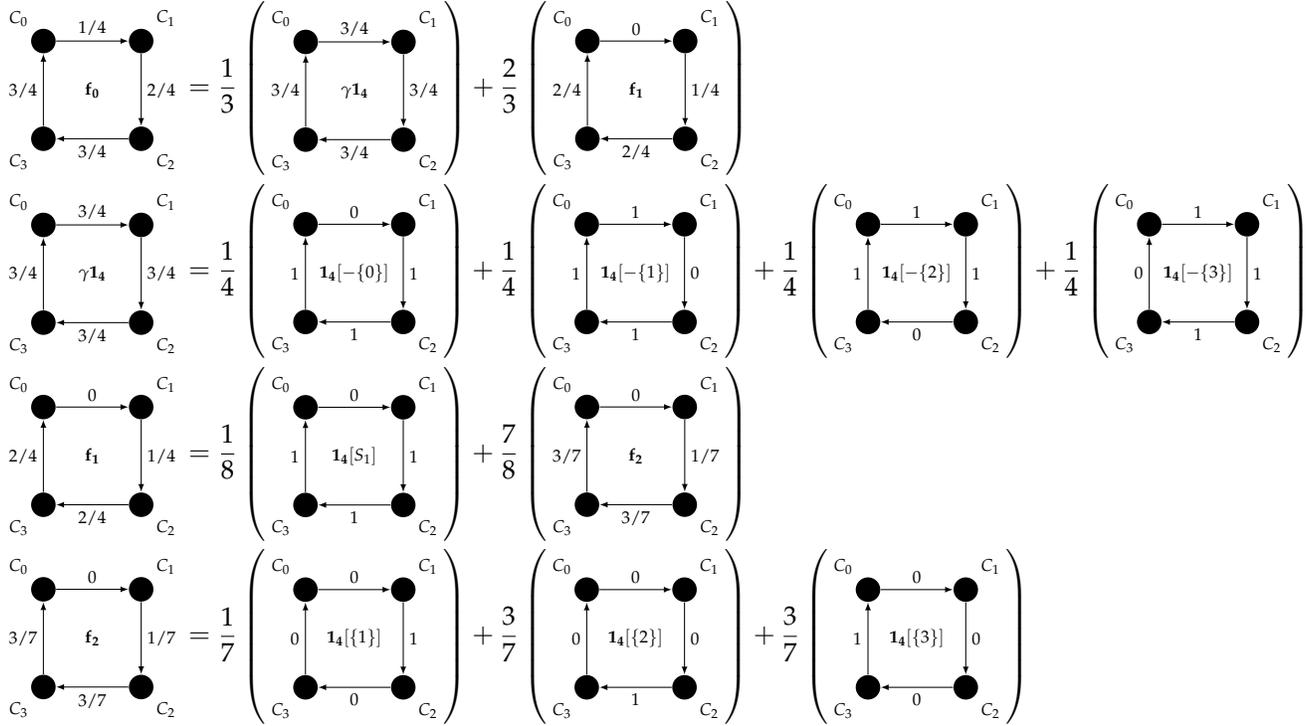

\centering
\begin{align*}
    \fourgraph{1/4}{2/4}{3/4}{3/4}{\vec{f_0}} &= \frac{1}{3} \left( \fourgraph{3/4}{3/4}{3/4}{3/4}{\gamma \vec{1_4}}\right) + \frac{2}{3} \left( \fourgraph{0}{1/4}{2/4}{2/4}{\vec{f_1}}\right)\\
    \fourgraph{3/4}{3/4}{3/4}{3/4}{\gamma \vec{1_4}} &= \frac{1}{4}\left(\fourgraph{0}{1}{1}{1}{\indvecc{-\{0\}}}\right) + \frac{1}{4}\left(\fourgraph{1}{0}{1}{1}{\indvecc{-\{1\}}}\right) + \frac{1}{4}\left(\fourgraph{1}{1}{0}{1}{\indvecc{-\{2\}}}\right) + \frac{1}{4}\left(\fourgraph{1}{1}{1}{0}{\indvecc{-\{3\}}}\right)\\
    \fourgraph{0}{1/4}{2/4}{2/4}{\vec{f_1}} &= \frac{1}{8}\left( \fourgraph{0}{1}{1}{1}{\indvecc{S_1}}\right) + \frac{7}{8}\left(\fourgraph{0}{1/7}{3/7}{3/7}{\vec{f_2}}\right)\\
    \fourgraph{0}{1/7}{3/7}{3/7}{\vec{f_2}} &= \frac{1}{7}\left(\fourgraph{0}{1}{0}{0}{\indvecc{\{1\}}}\right) + \frac{3}{7}\left(\fourgraph{0}{0}{1}{0}{\indvecc{\{2\}}}\right) + \frac{3}{7}\left(\fourgraph{0}{0}{0}{1}{\indvecc{\{3\}}}\right)
\end{align*}
\caption{Decomposing an expert graph $\vec{f_0}$ following the procedure given in Appendix Section~\ref{apx:decomposecycles}. The first line pulls off a scaled $\vec{1_4}$. The second line decomposes this by Observation~\ref{obs:decompose_cycle}. The third line decomposes $\vec{f_1}$ into $\indvecc{S_1}$ and $\vec{f_2}$ with $\onenorm{f_2} = 1$. The fourth line decomposes $\vec{f_2}$ according to Observation~\ref{obs:finishing_decomp}.}
\label{fig:knowledge_graph_decomp_algo}
\end{figure*}

\end{IEEEproof}
\end{document}